\documentclass[preprint,3p]{elsarticle}

\usepackage{amssymb}
\usepackage{amsmath}
\usepackage{dsfont}
\usepackage{algorithm}
\usepackage{algpseudocode}
\usepackage{amsmath, dsfont}
\usepackage{amssymb}
\usepackage{amsfonts}
\usepackage{amsthm}
\usepackage{amscd}
\usepackage{multirow}
\usepackage{appendix}
\usepackage{bbm}
\usepackage{braket}
\usepackage{bookmark}
\usepackage{booktabs}
\usepackage{caption}
\usepackage{cases}
\usepackage{enumitem}
\usepackage[T1]{fontenc} 
\usepackage{float}
\usepackage{lineno,hyperref}
\modulolinenumbers[5]
\usepackage{indentfirst}
\usepackage[utf8x]{inputenc}
\usepackage{verbatim}
\usepackage[autostyle,italian=guillemets]{csquotes}
\usepackage{graphicx}
\graphicspath{{Images/}}
\DeclareGraphicsExtensions{{.png},{.pdf},{.jpg}}
\graphicspath{{./}}
\usepackage{lmodern}
\usepackage{lipsum} 
\usepackage{listings}
\usepackage{yhmath}
\usepackage[table]{xcolor}
\usepackage{mathtools}
\usepackage{mathrsfs}

\usepackage{microtype} 
\usepackage{multirow}
\usepackage{setspace}
\usepackage{standalone}
\usepackage{subcaption}
\usepackage{tabularx}
\usepackage[table]{xcolor}
\usepackage{wasysym}
\usepackage{wrapfig}
\usepackage{url} 
\usepackage{wrapfig}
\usepackage{lscape}
\usepackage{rotating}
\usepackage{siunitx}
\usepackage[textsize=footnotesize]{todonotes}
\usepackage[overload]{empheq}
\usepackage{cleveref} 
\usepackage[acronym]{glossaries}

\makeglossaries

\newacronym{fcn}{FCN}{Fully Convolutional Network}
\newacronym{resnet}{ResNet}{}
\newacronym{ecg}{ECG}{electrocardiogram}
\newacronym{lv}{LV}{left ventricle}
\newacronym{rv}{RV}{right ventricle}
\newacronym{mv}{MV}{mitral valve}
\newacronym{tv}{TV}{tricuspid valve}
\newacronym{gradcam}{Grad-CAM}{}
\newacronym{ggradcam}{Guided Grad-CAM}{}
\newacronym{guidedback}{Guided Backpropagation}{}
\newacronym{wpw}{WPW}{Wolff-Parkinson-White}
\newacronym{xai}{XAI}{eXplainable Artificial Intelligence}
\newacronym{cnn}{CNN}{Convolutional Neural Network}
\newacronym{ap}{AP}{accessory pathway}
\newacronym{avrt}{AVRT}{atrioventricular re-entry tachycardia}
\newacronym{av}{AV}{atrioventricular}
\newacronym{uvc}{UVC}{universal ventricular coordinate}
\newacronym{qrs}{QRS}{}
\newacronym{gap}{GAP}{Global Average Pooling}
\newacronym{dl}{DL}{Deep Learning}
\newacronym{nn}{NN}{Neural Network}
\newacronym{dt}{DT}{diagnostic tree}
\newacronym{dtw}{DTW}{Dynamic Time Warping}
\newacronym{ml}{ML}{machine learning}
\newacronym{ep}{EP}{electrophysiology}

\definecolor{LightGreen}{RGB}{200,255,200}
\definecolor{LightYellow}{RGB}{255,255,180}

\journal{Medical Image Analysis}

\begin{document}

\begin{frontmatter}



\title{Explainable Deep Learning-based Classification of Wolff-Parkinson-White Electrocardiographic Signals}

\author[1,2]{Alice Ragonesi} 
\ead{ragonesi@uw.edu}
\author[1]{Stefania Fresca}
\ead{sfresca@uw.edu}
\author[2,3,4]{Karli Gillette}
\ead{karli.gillette@medunigraz.at}
\author[2]{Stefan Kurath-Koller}
\ead{stefan.kurath@medunigraz.at}
\author[2,5]{Gernot Plank}
\ead{gernot.plank@medunigraz.at}
\author[2,5]{Elena Zappon\corref{cor1}}
\ead{elena.zappon@medunigraz.at}
\cortext[cor1]{Corresponding author}
\affiliation[1]{organization={Department of Mechanical Engineering, University of Washington},
            city={Seattle},
            state={WA},
            country={USA}}
\affiliation[2]{organization={Division of Biophysics and Medical Physics, Gottfried Schatz Research Center, Medical University of Graz},
city={Graz},
country={Austria}}
\affiliation[3]{organization={Scientific Computing and Imaging Institute, University of Utah},
city={SLC},
state={UT},
country={USA}}
\affiliation[4]{organization={Department of Biomedical Engineering, University of Utah},
city={SLC},
state={UT},
country={USA}}
\affiliation[5]{organization={BioTechMed-Graz},
city={Graz},
country={Austria}}

\begin{abstract}
\gls{wpw} syndrome is a cardiac \gls{ep} disorder caused by the presence of an \gls{ap} that bypasses the atrioventricular node, faster ventricular activation rate, and provides a substrate for atrio-ventricular reentrant tachycardia (AVRT). Accurate localization of the \gls{ap} is critical for planning and guiding catheter ablation procedures. While traditional \gls{dt} methods and more recent \gls{ml} approaches have been proposed to predict \gls{ap} location from surface \gls{ecg}, they are often constrained by limited anatomical localization resolution, poor interpretability, and the use of small clinical datasets.

In this study, we present a \gls{dl} model for the localization of single manifest \glspl{ap} across 24 cardiac regions, trained on a large, physiologically realistic database of synthetic \glspl{ecg} generated using a personalized virtual heart model. 
We also integrate \gls{xai} methods, \gls{guidedback}, \gls{gradcam}, and \gls{ggradcam}, into the pipeline. This enables interpretation of \gls{dl} decision-making and addresses one of the main barriers to clinical adoption: lack of transparency in \gls{ml} predictions.

Our model achieves localization accuracy above 95\%, with a sensitivity of 94.32\% and specificity of 99.78\%. \gls{xai} outputs are physiologically validated against known depolarization patterns, and a novel index is introduced to identify the most informative \gls{ecg} leads for \gls{ap} localization. Results highlight lead V2 as the most critical, followed by aVF, V1, and aVL. This work demonstrates the potential of combining cardiac digital twins with explainable \gls{dl} to enable accurate, transparent, and non-invasive \gls{ap} localization.
\end{abstract}





\end{frontmatter}



\section{Introduction}
\gls{wpw} syndrome is a heart rhythm disorder characterized by the presence of at least one \gls{ap} between the atria and the ventricles that bypasses the atrioventricular node \cite{vuatuașescu2024wolf}. These \glspl{ap} enable conduction outside the normal nodal pathway, which can facilitate abnormally fast ventricular rates, as atrial arrhythmias are transmitted directly to the ventricles without the filtering effect of the atrioventricular node. When antegradely conducting, \glspl{ap} cause ventricular pre-excitation \cite{sapra2020wolff}, altering the normal sequence of ventricular activation. Furthermore, the presence of an \gls{ap} provides a substrate for atrioventricular reentrant tachycardia, which may occur through orthodromic or antidromic conduction circuits, producing narrow or wide QRS complexes, respectively.
\gls{wpw} affects up to 0.5\% of the general population \cite{elendu2025risk,paerregaard2023wolff}, with possible symptoms such as paroxysmal supraventricular tachycardia, atrial fibrillation, atrial flutter, ventricular fibrillation, and, eventually, sudden death depending on the manifestation and location of the pathways \cite{sapra2020wolff,schiavone2024pre,pappone2012risk}. Radiofrequency catheter ablation is the first-line therapy, aiming to eliminate the \gls{ap}, re-establish a non-pre-excited sinus rhythm, and impede the formation of \glspl{avrt}. However, approximately 6\% of ablation procedures remain unsuccessful \cite{sherdia2023success,fujino2020clinical}.

Accurate localization of the \gls{ap} prior to ablation is crucial for optimal procedural planning, reducing time for intervention and associated risks. Traditionally, single manifest \gls{ap} localization relies on analysis of the surface \gls{ecg}, either acquired during resting sinus rhythm \cite{lindsay1987concordance,milstein1987algorithm,fitzpatrick1994new,xie1994localization,chiang1995accurate,d1995fast,iturralde1996new,arrudaDevelopmentValidationECG1998,boersma2002accessory,kurath2022accuracy,elhamritiEASYWPWNovelECGalgorithm2023,khalaph2025novel} or under atrial pacing \cite{kurath2022accuracy,fananapazirImportancePreexcitedQRS1990,pambrunMaximalPreExcitationBased2018}. Characteristic \gls{ecg} features in \gls{wpw} syndrome with antegrade \glspl{ap} include a short PR interval, the presence of a delta wave, and a wide QRS complex. Localization algorithms, typically structured as diagnostic decision trees, associate these morphological features in selected leads with predefined anatomical regions along the \gls{mv} or \gls{tv} annulus. For instance, Arruda \emph{et al.}~\cite{arrudaDevelopmentValidationECG1998} proposed a diagnostic algorithm based on the polarity of the delta wave in leads I and V1 to determine whether the \gls{ap} is located along the \gls{mv} or \gls{tv}. To further refine localization, anteroseptal, posteroseptal, or lateral, they assessed the positivity, negativity, or isoelectricity of leads II, aVF, V1, and I. In contrast, more recent algorithms, such as those in Pambrun \emph{et al.}~\cite{pambrunMaximalPreExcitationBased2018}, El Hamriti \emph{et al.}~\cite{elhamritiEASYWPWNovelECGalgorithm2023}, and Khalaph \emph{et al.}~\cite{khalaph2025novel}, relied on the overall QRS complex polarity in lead V1 to determine left versus right annular location. These latter algorithms then used the polarity of the full QRS complex, rather than the delta wave alone, in inferior leads (II, III, aVF) and sometimes counted the number of positive and negative deflections to classify the \gls{ap} as anterolateral, posterolateral, or septal. However, a study employing a computational model for simulating \gls{wpw} syndrome with single manifest antegrade \gls{ap}, Gillette \emph{et al.}~\cite{gilletteComputationalStudyInfluence2025} showed that the largest variations in QRS morphology and delta wave due to the presence of an \gls{ap} were found in lead V2 for both \gls{lv} and \gls{rv}, while a very similar QRS complex was obtained in lead V1, especially for antegrade \gls{ap} located in the septum. Furthermore, manual classification of \gls{ap} location based on \glspl{dt} can be complex, time-consuming, and highly prone to inter-observer variability in identifying \gls{ecg} patterns \cite{mcgavigan2007localization,teixeira2016accuracy}, leading to an overall low predictive accuracy. Consequently, the development of rapid, automated, objective and non-invasive approaches to analyze surface \gls{ecg} signals is of growing interest to improve the identification of \glspl{ap} before ablation. 

\gls{ml} techniques have previously been applied to classify various cardiac pathologies based on morphological features extracted from the \glspl{ecg} \cite{jahmunahExplainableDetectionMyocardial2022,jonesImprovingECGClassification2020,vandeleurDiscoveringVisualizingDiseaseSpecific2021, leLightX3ECGLightweightEXplainable2022, vandeleurECGonlyExplainableDeep2024}. Unlike traditional approaches that rely on manually derived features, modern \gls{ml} methods, particularly \gls{dl}, enable automated extraction and identification of patterns within complex signals \cite{van2024automatic}. This automated analysis improves efficiency, reduces subjectivity, and significantly decreases the time and effort required for interpretation \cite{senonerIdentifyingLocationAccessory2021,nishimoriAccessoryPathwayAnalysis2021,de2023machine,van2024automatic,yahyazadehNovelFeatureExtraction2024,hennecken2025localization}. \gls{ml}-based algorithms for the classification of single manifest \glspl{ap} had been proposed in a number of works. For example, Yahyazadeh \emph{et al.}~\cite{yahyazadehNovelFeatureExtraction2024} implemented an automatic feature extraction based on the McSharry model to derive quantitative parameters from \gls{ecg} traces of 31 patients with \gls{wpw}, focusing on the polarity of the delta wave and the morphology of QRS. A random forest and a gradient boosting approach were then employed to assign the obtained features to either right or left ventricles \gls{ap} location, achieving a better accuracy compared to traditional \glspl{dt}. A combination of manual and automated methods for extracting \gls{ecg} features, such as delta wave polarity and QRS morphology, was employed by Bernardo \emph{et al.}~\cite{de2023machine}. These features were subsequently fed to random forest and gradient boosting algorithms to classify the \gls{ap} location in 120 patients as left, septal, or right. \gls{dl} techniques have been deployed in more recent studies \cite{senonerIdentifyingLocationAccessory2021,nishimoriAccessoryPathwayAnalysis2021,hennecken2025localization}. Senoner \emph{et al.}~\cite{senonerIdentifyingLocationAccessory2021} utilized a neural network with back propagation to classify \glspl{ap} in 107 patients across 14 distinct anatomical locations. Nishimori \emph{et al.}~\cite{nishimoriAccessoryPathwayAnalysis2021} developed a hybrid model combining \gls{ecg} signals with a \gls{cnn}-based encoding of chest X-ray images, feeding these into a \gls{cnn} to classify 206 \glspl{ap} into three anatomical groups. More recently, Hennecken \emph{et al.}~\cite{hennecken2025localization} applied a causal neural network with a multi-task learning framework to localize \glspl{ap} in 645 patients, categorizing them as right-sided, septal, or left-sided.

    Independent of the deployed technique, either \gls{ml} methods or traditional \glspl{dt} generally demonstrate acceptable accuracy in identifying lateral \gls{ap} positions along the \gls{tv} and \gls{mv} annuli. However, their accuracy in predicting septal \glspl{ap} remains limited. Most existing \glspl{dt} are designed to determine the circumferential (rotational) location of the \gls{ap}, i.e., assigning it to broad regions encircling the \gls{mv} or \gls{tv}. These models, however, fail to consider the possible longitudinal (apex-to-base) position of the \gls{ap}, which would allow for more precise localization of the \gls{ap} along the \gls{lv} and \gls{rv}. As a result, current algorithms rely on a coarse subdivision into a limited number of regions defined circumferentially around the valve annuli, substantially restricting their ability to pinpoint the \gls{ap} site, especially along the apico-basal direction. Additionally, despite the complexity of the \gls{dt}, the accuracy of these algorithms tends to be lower in pediatric populations \cite{boersma2002accessory,wren2012accuracy,li2019novel,baek2020new}. Recently, Khalaph \emph{et al.} \cite{khalaph2025novel} demonstrated that a simplified \gls{dt} can reach an overall accuracy of 97\% in both adult and children. Nevertheless, it still struggles with more complex cases involving multiple \glspl{ap} or coexisting structural heart disease. 

\gls{ml}-based techniques have generally demonstrated superior accuracy compared to traditional \gls{dt} approaches for localizing accessory pathways. However, the performance of these models is often constrained by the limited size of the training datasets, which significantly affects both the robustness and generalizability of the results. The availability of large, high-quality annotated datasets is critical for the effective training and validation of \gls{ml} models. However, acquiring such datasets, particularly in the context of rare or heterogeneous \gls{ep} conditions, remains a key limitation, hindering the full potential of data-driven approaches in this domain. Furthermore, \gls{ml} techniques are considered black-box models, since their internal decision-making processes lack transparency, and can thus undermine clinical trust and diminish their acceptance in medical practice \cite{siontisSaliencyMapsProvide2023, ayanoInterpretableMachineLearning2022}.

To address the limitations posed by small clinical datasets, virtual models of cardiac \gls{ep} that accurately replicate activation and repolarization in patients offer a compelling solution. These models enable the simulation of all cardiac activation patterns relevant in \gls{wpw} patients, including anti- and orthodromic \glspl{avrt}, such as those observed in \gls{wpw} syndrome, in a controlled, reproducible, and scalable manner \cite{loewe2022cardiac,corral2020digital}. When personalized in terms of both cardiac anatomy and \gls{ep} properties, often referred to as cardiac digital twins, these models can be leveraged for diagnostic refinement, prognostic evaluation, and treatment planning \cite{dasi2024silico,bhagirath2024bits,qian2025developing}. Furthermore, they provide a valuable platform for generating synthetic data to supplement limited clinical datasets, enhancing the training and reliability of \gls{ml} algorithms \cite{doste2022training,gillette2023medalcare}. In this context, Gillette \emph{et al.}~\cite{gilletteComputationalStudyInfluence2025} developed a subject-specific computational model of \gls{wpw} syndrome to investigate how variations in \gls{ap} location influence \gls{ecg} morphology, by creating a larger synthetic cohort of \glspl{ecg} compared to the ones previously used in \gls{ap} localization works. 

 To mitigate concerns over transparencyency and reliability, \gls{xai} methods \cite{samekExplainableArtificialIntelligence2019} have been developed and applied in \gls{ecg} classification tasks, helping to clarify how predictions are made. However, to the best of our knowledge, no study has so far implemented these techniques in the context of \gls{wpw} syndrome. Incorporating \gls{xai} into \gls{ml} models for \gls{ap} localization would greatly enhance transparency, physician confidence, and clinical applicability \cite{jonesImprovingECGClassification2020,vandeleurDiscoveringVisualizingDiseaseSpecific2021,jahmunahExplainableDetectionMyocardial2022,leLightX3ECGLightweightEXplainable2022, vandeleurECGonlyExplainableDeep2024}.

Building on the virtual database of antegrade single manifest \glspl{ap} introduced in \cite{gilletteComputationalStudyInfluence2025}, this work presents a \gls{dl}-based model for the localization of \glspl{ap} across 24 distinct cardiac regions. The model is integrated with \gls{xai} techniques to enhance transparency and explainability of the \gls{dl} inference process. Using a calibrated digital twin of a single subject in sinus rhythm, the simulated \glspl{ap} produce significant variation in \gls{ecg} morphology, spanning a broad and physiologically relevant range of possible \gls{ap} locations. This results in a rich and comprehensive dataset of 8842 12-lead \glspl{ecg} suitable for learning the relationship between \gls{ap} position and surface \gls{ecg} features.
The 24 selected regions account for both rotational and longitudinal positioning of the \gls{ap}, thereby enabling more precise localization. To explore optimal input data dimensionality to maximize the network feature extraction capability, we tested three different \gls{cnn} architectures. To select the best \gls{xai} method to define model transparency, we implemented and compared three \gls{xai} techniques for \gls{ecg} classification interpretation in the form of \gls{guidedback} \cite{springenbergStrivingSimplicityAll2015}, \gls{gradcam}, and \gls{ggradcam} \cite{selvarajuGradCAMVisualExplanations2020}. Although initially developed for image data, these methods have recently demonstrated strong potential in identifying \gls{ecg} morphological features associated with specific cardiac pathologies \cite{jonesImprovingECGClassification2020,vandeleurDiscoveringVisualizingDiseaseSpecific2021,jahmunahExplainableDetectionMyocardial2022,leLightX3ECGLightweightEXplainable2022, vandeleurECGonlyExplainableDeep2024}.

The evaluation of our novel \gls{dl}-based method on a high-fidelity synthetic dataset with known ground truth showed better performance than traditional \glspl{dt} and existing \gls{ml} approaches, achieving localization accuracy above 95\%, with a sensitivity of 94.32\% and specificity of 99.78\% across the 24 regions. %
Furthermore, we clustered similar \gls{xai} outputs within each region, and correlated such results with \gls{ep} observations of depolarization patterns caused by \glspl{ap}, providing physiological validation of the model decision-making process. Finally, we introduce a novel index that summarizes \gls{xai} results to identify the most informative leads for \gls{ap} detection in the left and right ventricles. Our findings are consistent with those reported in \cite{gilletteComputationalStudyInfluence2025}, showing that lead V2 contributes the most to localization accuracy, followed by aVF, V1, and aVL, with lead I being the least informative.

The paper is organized as follows: Section~\ref{Sec:methods} presents the overall methodology, including the generation of the dataset (Section~\ref{Subsec:dataset}), the \gls{cnn} architectures (Section~\ref{Subsec:CNNs}), and the \gls{xai} techniques (Section~\ref{subsec:explainability}). Section~\ref{Sec:results} reports the results, including the accuracy of the proposed methodology (Section~\ref{subsec:classification}), the physiological interpretation of the \gls{xai} outputs (Section~\ref{subsec:explainability_results}), and the proposed index for summarizing \gls{xai} relevance (Section~\ref{subsec:important_leads}). The discussion, limitations, and conclusions are provided in Sections~\ref{sec:discussion}, \ref{sec:limitaions}, and \ref{sec:conclusions}, respectively.

\section{Methods}
\label{Sec:methods}
In this section, we describe the \gls{dl} models employed for the \gls{ecg} classification, together with the training and testing datasets. Furthermore, the explored \gls{xai} methods for the interpretation of the \gls{dl} decision-making process are discussed. Overall, three \glspl{fcn} were trained, validated, and tested on synthetic \gls{ecg} data. For interpretability, only the network with 2D kernels was retained, as it allowed the computation of lead-specific feature relevance using \gls{ggradcam} saliency maps. Explainability analysis was performed exclusively on correctly classified test samples. \gls{xai} results were evaluated both globally - across all samples, to define a lead importance index —and locally - through saliency maps of representative \glspl{ecg} — to assess the alignment between model attention and physiologically meaningful features.
 
\subsection{Dataset}
\label{Subsec:dataset}
\begin{figure}[!t]
    \centering    \includegraphics[width=0.7\linewidth, keepaspectratio]{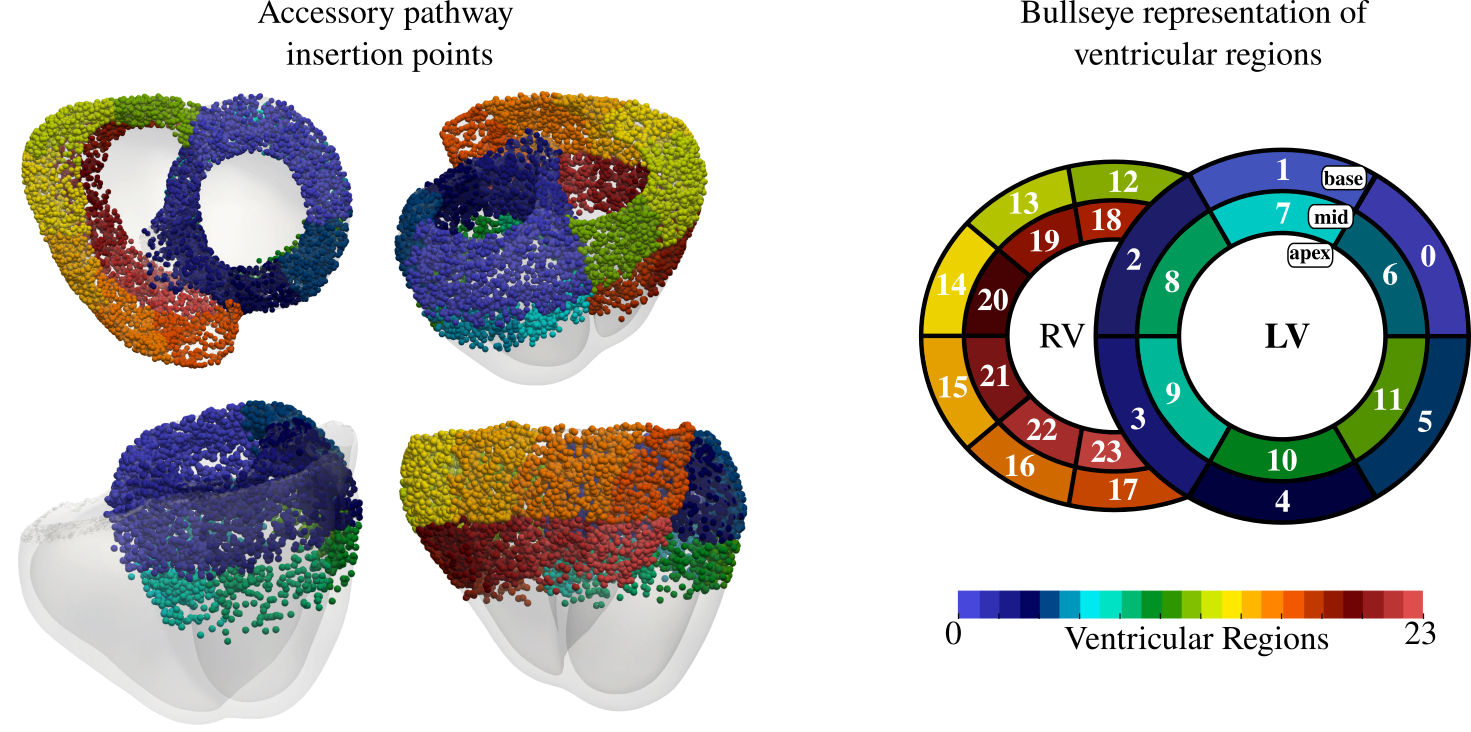}
    \caption{\gls{ap} insertion points grouped according to the defined regions (left) and bullseye representation of the anatomical region distribution (right). The regions were defined based on \gls{uvc}, including both rotational and longitudinal subdivision.}
\label{fig:regions_visualization}
\end{figure}

\begin{figure}[!t]
    \centering
    \includegraphics[width=0.9\linewidth]{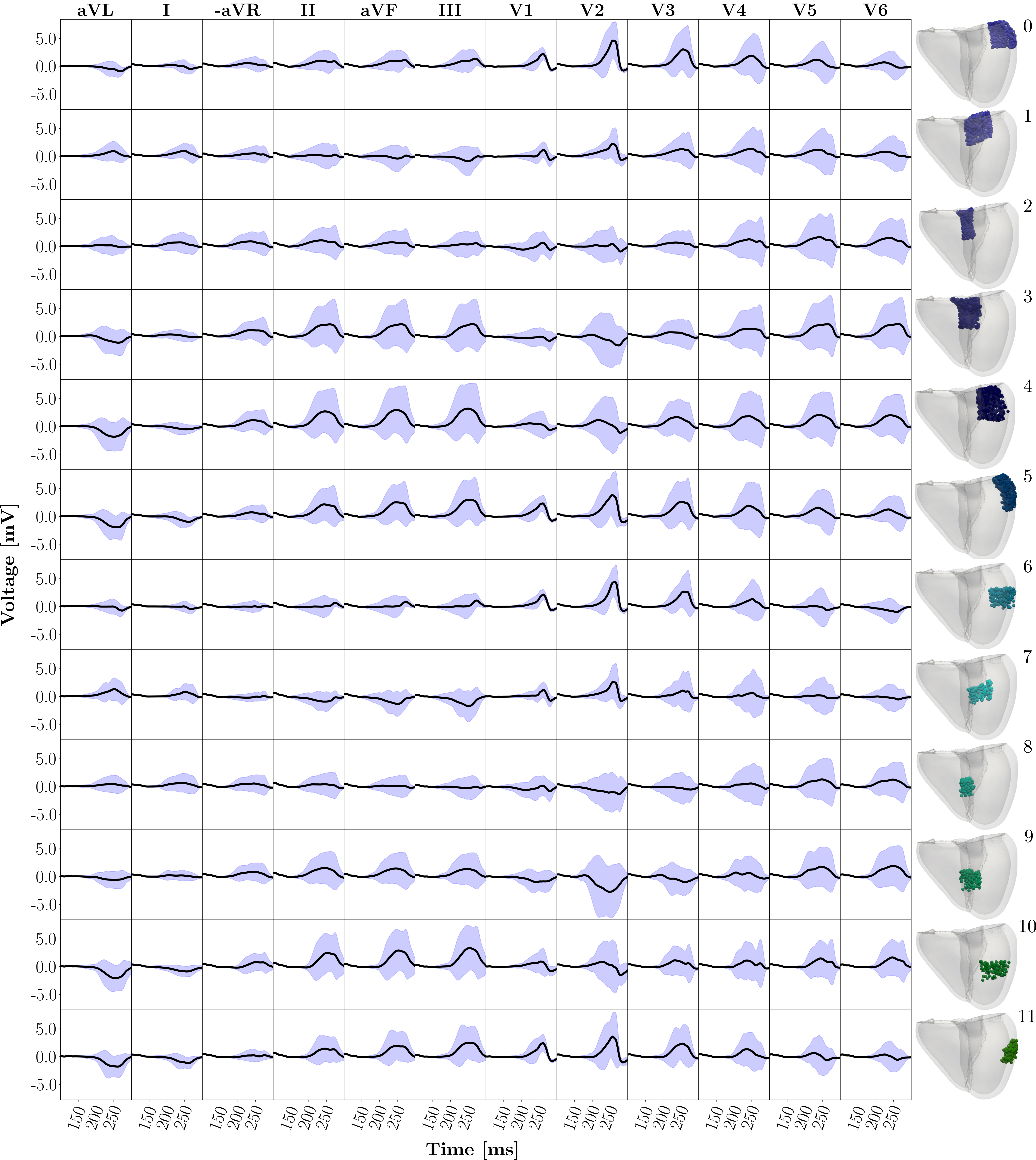}
    \caption{Regional variation of the signals with \gls{ap} in the \gls{lv}. Mean signal (black) with shaded regions representing plus or minus 2 standard deviations across samples. Next to the leads, the \glspl{ap} insertion sites in the ventricles are displayed together with the region numbering.}
    \label{fig:lv}
\end{figure}

\begin{figure}[!t]
    \centering
    \includegraphics[width=0.9\linewidth]{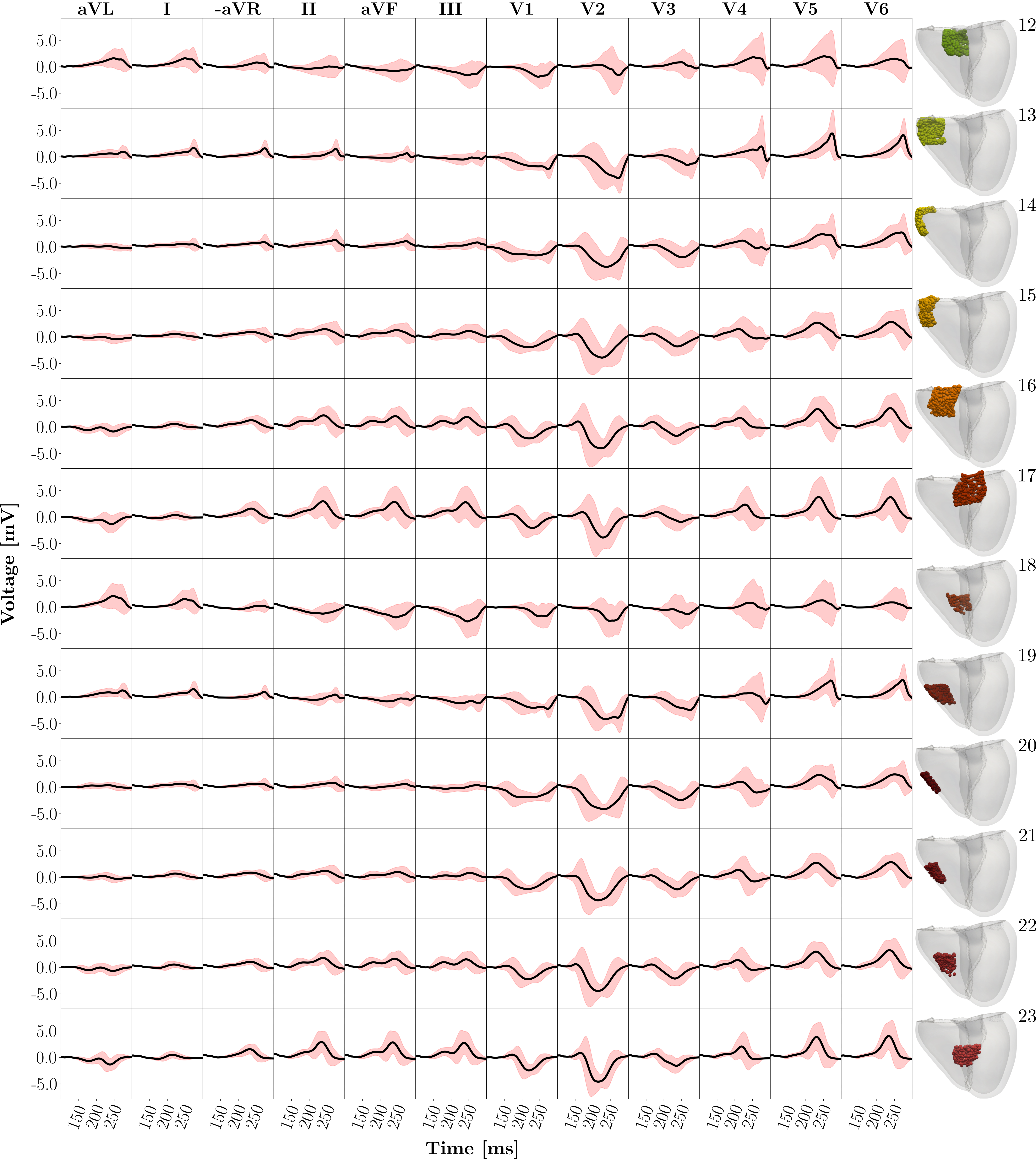}
    \caption{Regional variation of the signals with \gls{ap} in the \gls{rv}. Mean signal (black) with shaded regions representing plus or minus 2 standard deviations across samples. Next to the leads, the \glspl{ap} insertion sites in the ventricles are displayed together with the region numbering.}
    \label{fig:rv}
\end{figure}

The dataset used in this work consists of the \emph{in silico} 12-lead pathological \glspl{ecg} dataset generated in \cite{gilletteComputationalStudyInfluence2025}. A cardiac digital twin of one healthy subject was created, comprising a four-chamber heart and the corresponding torso, and calibrated using the subject’s 12-lead \gls{ecg}. A set of \glspl{ap} was subsequently generated based on physiological constraints. Each \gls{ap} entry site in the ventricular myocardium was expressed through the rotational ($\phi$), apico-basal ($z$) and transmural ($\rho$) \glspl{uvc}. The \glspl{ap}, along with the corresponding \glspl{ecg}, were grouped into 12 regions per ventricle, similar to the AHA model, resulting in $C = 24$ regions serving as the \glspl{ecg} labels for the classification task (see Figure \ref{fig:regions_visualization}). The rotational and apico-basal \glspl{uvc} were used as reference cutoffs to delineate the anatomical regions for AP localization. The rotational coordinate sliced the ventricles around the mitral and tricuspid valve, while the apico-basal coordinate bisected each slice. The septal \glspl{ap} were attributed to regions in the \gls{lv}. In total, the dataset comprised $N = 8842$ \glspl{ecg}, with 4506 associated to \glspl{ap} in the \gls{lv} and 4338 to \glspl{ap} in the \gls{rv}. The \gls{ecg} time series was restricted to the $[100,300]$ \SI{}{\milli\second} interval, comprising $T=200$ time steps, corresponding to the portion of the QRS complex. Indeed, morphological features in the QRS complex have been physiologically linked to \gls{wpw} syndrome \cite{fananapazirImportancePreexcitedQRS1990, haissaguerreElectrocardiographicCharacteristicsCatheter1994}. During this interval, R peaks were not aligned, as the timing of pre-excitation, while shifting signals along the time axis, could potentially be a characteristic upon which a classifier learns to assign samples to regions. Anatomical regions and data variability associated to each region can be seen in Figures \ref{fig:regions_visualization}-\ref{fig:rv}.

From a mathematical standpoint, 12-lead \glspl{ecg} are multivariate time series with $L = 12$ channels, or dimensions, each representing the voltage in a lead over time. Each instance can be easily treated as a matrix $\mathbf{S} = [\mathbf{s}_1;\dots; \mathbf{s}_L]\in\mathbb{R}^{T\times L}$, where $\mathbf{s}_i\in\mathbb{R}^T$, $i=1,\dots,L$, stores the univariate time series associated to a single lead. Other representations of the \gls{ecg} signals are possible by reshaping $\mathbf{S}$, and were explored in this paper to learn multiple classifiers. Thus, in the following, $\mathbf{S}_r$ will refer to the reshaped version of matrix $\mathbf{S}$.

\subsection{Architectures}
\label{Subsec:CNNs}
Multiple nonlinear \gls{fcn} \cite{fawazDeepLearningTime2019} classifiers $\mathcal{G}(\mathbf{S}
_{r}; \mathbf{\Omega}): \mathbb{R}^{D} \rightarrow [0, 1]^C$ were trained to predict the probabilities $\mathbf{\hat{y}} = (\hat{y}_0, \dots, \hat{y}_{C-1})^T\in\mathbb{R}^C$  of a signal $\mathbf{S}_r$, formatted as to have dimensionality $D$, of belonging to each class $c = 0, \dots, C-1$, based on an optimal set of weights $\mathbf{\Omega}$. The core structure common to all \gls{fcn} models consisted of $\mathcal{L} = 3$ consecutive convolutional blocks, each comprising a convolutional layer followed by a BatchNormalization for regularization \cite{ioffeBatchNormalizationAccelerating2015} and a ReLU activation \cite{glorotDeepSparseRectifier2011}. The final block preceded a \gls{gap} layer \cite{linNetworkNetwork2014}, whose output was passed through a fully connected layer and a Softmax activation \cite{bishopPatternRecognitionMachine2006}. The convolutions were used to make the classes linearly separable and to filter the most important features of the signal $\mathbf{S}_r$ \cite{haykinNeuralNetworksLearning2009a}. In the $i$-th convolutional block, $i =1,\dots,\mathcal{L}$, $M_i$ convolutional filters are swept over the input $\mathbf{X}_{i-1}$, with stride $s_i$ controlling the step size at which the filters move across the input. After the BatchNormalization $\mathcal{BN}$ and ReLU transformations, a multivariate feature map $\mathbf{X}_{i} = [\mathbf{x}_i^1;\dots; \mathbf{x}_i^{M_i}]\in\mathbb{R}^{D_i \times M_i}$ with $M_i$ channels is provided as output, such that each channel $m = 1,\dots,M_{i}$ holds the following nonlinear representation of the input:
\begin{equation}\label{eq:conv}
    \mathbf{x}_{i}^{m}  = \mathrm{ReLU}\bigg(\mathcal{BN}\bigg(\sum_{j=1}^{M_{i-1}}\mathbf{w}_{i,j}^m *_{s_i} \mathbf{x}_{i-1}^{j}\bigg)\bigg),
\end{equation}
where $*_{s_i}$ denotes the convolution operation with stride $s_i$, $\mathbf{w}_{i,j}^m$ is the set of weights defining the $j$-th kernel of filter $m$ in the $i$-th convolutional layer, and $\mathbf{X}_0 = \mathbf{S}_r$. The dimensionality $D_i$ of the features obtained for each channel is affected by the input's shape, as discussed in the following in Sections \ref{subsec:stack}-\ref{subsec:image}.

After the nonlinear transformations, the \gls{gap} layer is used to flatten  the multivariate feature map $\mathbf{X}_{\mathcal{L}}$ into the vector $\mathbf{v}\in\mathbb{R}^{M_\mathcal{L}}$ collecting the average of each channel, then linearly mapped by the fully connected layer to the vector of logits $\mathbf{z} = (z_0,\dots,z_{C-1})\in\mathbb{R}^C$. Finally, the Softmax activation function is employed to perform the classification, that is, to compute the probabilities $\hat{y}_c$, $c=0,\dots,C-1$, such that
\begin{equation*}
    \hat{y}_c = \frac{e^{z_c}}{\sum_{j=0}^{C-1}e^{z_j}}.
\end{equation*}

The training was performed on the training set $I_{train}$, employing a gradient descent algorithm to minimize the cross-entropy loss function
\begin{equation*}
    CCE(\mathbf{y}, \hat{\mathbf{y}}) = - \sum_{c=0}^{C-1}y_{c}\log{(\hat{y}_c)}.
\end{equation*}
between the ground truth $\mathbf{y}$ and predicted $\hat{\mathbf{y}}$, where $\mathbf{y}$ is a one-hot-encoded label vector $\mathbf{y} = (y_0, \dots, y_{C-1})\in\mathbb{R}^C$, $y_c \in \{0, 1\}^C$.

In this work, the core \gls{fcn} architecture was kept unchanged while evaluating its classification performance under different input configurations, leading to three distinct \gls{fcn} models. Specifically, the \gls{ecg} was represented as:
(i) a univariate time series formed by sequentially stacking all leads, resulting in a series of length $L \cdot T$ (Section \ref{subsec:stack});
(ii) a multi-dimensional signal comprising $L$ separate one-dimensional time series (Section \ref{subsec:multichannel}); and
(iii) a two-dimensional image with height $T$ and width $L$ (Section \ref{subsec:image}).
Indeed, variations in input dimensionality called for differences in the neural network architectures, in terms of the number of channels at the input layer and kernel dimensionality.

\subsubsection{Single-channel \gls{fcn} with 1D kernels}
\label{subsec:stack}
One strategy for processing multivariate time series is to reshape each data sample $\mathbf{S}$ into a longer univariate sequence $\mathbf{S}_r=[\mathbf{s}_1^T;\dots; \mathbf{s}_{L}^T]^T \in \mathbb{R}^{T\cdot L}$ by stacking the individual dimensions end-to-end. In such a setting, the convolutional layers are constituted of 1D convolutional kernels sliding over the flattened sequence. The kernel length must therefore be carefully set to capture dependencies both within individual dimensions (intra-dimensional) and across different dimensions (inter-dimensional). 

In this configuration, the 12 leads were stacked to represent each \gls{ecg} as a univariate sequence of 2400 time steps, which was then processed by a \gls{fcn} with a single input channel. For each convolutional layer, the number of kernels, kernel size, and stride were treated as tunable hyperparameters. Given the absence of prior knowledge regarding whether the most informative features for classification resided within individual leads or across inter-lead relationships, the design prioritized the extraction of pathology-relevant features at the lead scale. It was hypothesized that employing kernel sizes comparable to the full length of the stacked leads could result in excessive information loss. Accordingly, kernel sizes approximately matching the length of a single lead were preferred, with values up to 100 permitted in the first convolutional layer. The stride parameter was also tuned, as maintaining a fixed stride of 1 -- as implemented in Sections~\ref{subsec:multichannel} and~\ref{subsec:image} -- could result in the extraction of redundant features due to the effects of the convolution and weight sharing (see Eq.~\eqref{eq:conv}), particularly when used in conjunction with larger kernels. Accordingly to input shape and network hyperparameters, the feature maps extracted by the convolutional layers were multivariate time series $\mathbf{X}_i\in\mathbb{R}^{T_i\times M_i}$, with length $T_i < T\cdot L$, $i = 1,\dots,\mathcal{L}$, as a consequence of the stride being larger than one. 


\subsubsection{Multi-channel FCN with 1D kernels}\label{subsec:multichannel}
As an alternative input formatting, each \gls{ecg} signal was arranged into a matrix, i.e. $\mathbf{S}_r = \mathbf{S}\in\mathbb{R}^{T\times L}$, which was fed into a \gls{fcn} with $L$ input channels -- one for each dimension of the multivariate time series, as in Rosafalco \emph{et al.} \cite{rosafalcoOnlineStructuralHealth2021}. In this configuration, a single 1D signal is fed to each input channel of the \gls{fcn}, thus requiring 1D kernels for performing the convolutions, with kernel lengths to be optimized through tuning operations. Here, 1D kernels with appropriate padding and a stride of 1 extract one dimensional feature maps, i.e. time series, with length equal to the input. Therefore, considering $M_i$ channels, the generated latent representations are $T$ steps long multivariate time series $\mathbf{X}_i\in\mathbb{R}^{T\times M_i}$.

Consequently to the formatting, the features extracted channel wise were combined through Eq.~\eqref{eq:conv} already at the first layer, thus allowing the network to optimize a summary representation of the features gathered from all leads.  

\subsubsection{Single-channel \gls{fcn} with 2D kernels}\label{subsec:image}
In the final approach to \gls{ecg} classification, the signals were treated as single-channel images, similarly to the method proposed by Ramírez \emph{et al.} \cite{ramirezArtSelectingECG2024}. Each signal was represented as a tensor $\mathbf{S}_r \in\mathbb{R}^{T\times L \times 1}$, where the height and width of the image corresponded to the signal length $T$ and the number of leads $L$, respectively. This representation necessitated the use of two-dimensional convolutions, wherein squared 2D kernels were employed to capture both inter-lead and intra-lead relationships, learning to represent them in 2D feature maps, i.e. $\mathbf{X}_i\in\mathbb{R}^{T\times L \times M_i}$. 

Treating \glspl{ecg} as images implied a one-to-one correspondence between image columns and ECG leads, which was subsequently leveraged to interpret the individual contributions of different leads to the classification outcomes, as discussed in Section \ref{subsec:explainability}.


\subsection{Comparison with established algorithms}\label{subsec:comparison}
The performance of our best model was evaluated against the \gls{dt} EASY-WPW algorithm proposed by El Hamriti and coauthors \cite{elhamritiEASYWPWNovelECGalgorithm2023}, as well as the \gls{dt} algorithm developed by Arruda and and coauthors  \cite{arrudaDevelopmentValidationECG1998}. To account for differences in anatomical subdivisions across reference regions, the \glspl{ecg} in our dataset were regrouped according to the \gls{dt} region definitions in \cite{elhamritiEASYWPWNovelECGalgorithm2023} and \cite{arrudaDevelopmentValidationECG1998}. The complete description of the new ground truth subdivision and of the overall comparison procedure is given in detail in \ref{sec:appendix_regions}. To compare the effectiveness of the \gls{fcn} against the \gls{dt} algorithms, a one-sided Fisher’s exact test was applied to the contingency table of correctly versus incorrectly classified samples. Statistical significance was determined using a threshold of $\alpha = 0.05$. 


\subsection{Network explainability}\label{subsec:explainability}

Interpreting deep learning models, especially \glspl{cnn}, remains a significant challenge due to their inherent “black-box” nature. Gaining insight into the features these models learn is essential for enhancing transparency and building trust in their predictions. In this work, we investigate post-hoc \gls{xai} techniques -- namely \gls{gradcam} \cite{selvarajuGradCAMVisualExplanations2020}, \gls{guidedback} \cite{springenbergStrivingSimplicityAll2015}, and their combination, \gls{ggradcam} \cite{selvarajuGradCAMVisualExplanations2020}.

Post-hoc \gls{xai} methods operate by attributing a relevance score to input features, effectively defining a scalar function that ranks, by ascending score values, the importance of each feature with respect to the classifier output \cite{theisslerExplainableAITime2022}. The resulting set of scores, termed a saliency map, explains the influence of input components on the computation of a specific model prediction, typically a logit $z_c$ of the last fully connected layer, associated with class $c$. 

Among the three investigated techniques, \gls{guidedback} produces high-resolution saliency maps by modifying the standard backpropagation, thus emphasizing fine-grained input features that strongly activate the network, while not inherently focusing on any specific class. In contrast, \gls{gradcam} generates class-discriminative saliency maps by computing gradients of the target class logit with respect to the final convolutional layer feature maps; these gradients are then pooled to produce coarse spatial importance maps highlighting the regions most relevant to the class prediction \cite{selvarajuGradCAMVisualExplanations2020}. Finally, \gls{ggradcam} combines the strengths of both methods, merging \gls{gradcam} class-discriminative localizations with \gls{guidedback} high-resolution details, thereby providing interpretable and precise visual explanations simultaneously \cite{selvarajuGradCAMVisualExplanations2020}.

In the following sections, we describe the \gls{guidedback}, \gls{gradcam} and \gls{ggradcam} computations, highlighting how the second poses a major limitation in the number of computed relevance scores depending on the structure of the \glspl{ecg} as model input data, and thus the model interpretability. Furthermore, a new metric termed \textit{lead importance} is introduced to assess which leads contain, more often than others, features that are ranked as important in the decision making process of the network. 



\subsubsection{Guided  Backpropagation}
\gls{guidedback} is a gradient-based approach to generate saliency maps that measure the relevance of each input feature in $\mathbf{S}_r$ in the computation of a target output score $z_c$. Specifically, the saliency map to interpret the logit $z_c$ obtained for an input signal $\mathbf{S}_r$ is defined as
\begin{equation}\label{eq:guidedb}
    \mathbf{B}_c^{\mathbf{S}_r} = \frac{\partial z_c}{\partial \mathbf{S}_r} = \frac{\partial z_c}{\partial \mathbf{X}_0},\:\: c = 0,\dots, C-1,
\end{equation}
where each gradient is computed by recursively applying the chain rule from the output to the input layer, modifying the standard backpropagation computations.
Indeed, in standard backpropagation, the gradient with respect to the feature map $\mathbf{X}_{i-1}$ -- i.e., one factor in the chain rule -- is given by
\begin{equation*}
    \frac{\partial z_c}{\partial \mathbf{X}_{i-1}} = \frac{\partial \mathbf{X}_i}{\partial \mathbf{X}_{i-1}} \cdot \frac{\partial z_c}{\partial \mathbf{X}_{i}},
\end{equation*}
which is modified in \gls{guidedback} into
\begin{equation}\label{eq:modified_backprop}
\frac{\partial z_c}{\partial \mathbf{X}_{i-1}} =
\left\{
\begin{array}{ll}
\displaystyle
\frac{\partial \mathbf{X}_i}{\partial \mathbf{X}_{i-1}} \cdot \frac{\partial z_c}{\partial \mathbf{X}_{i}} \cdot \mathds{1}\left( \frac{\partial z_c}{\partial \mathbf{X}_i} > 0 \right)
& \text{if layer $i$ is a ReLU}, \\[12pt]
\displaystyle
\frac{\partial \mathbf{X}_i}{\partial \mathbf{X}_{i-1}} \cdot \frac{\partial z_c}{\partial \mathbf{X}_{i}}
& \text{otherwise},
\end{array}
\right.
\end{equation}
where $\cdot$ denotes the element-wise product. The gradient of the ReLU-activated output $\mathbf{X}_i$ with respect to the input $\mathbf{X}_{i-1}$ is
\begin{equation*}
    \frac{\partial \mathbf{X}_i}{\partial \mathbf{X}_{i-1}} = \mathds{1}\left(\mathbf{X}_{i-1} > 0\right),
\end{equation*}
which holds for both standard backpropagation and its modified version. In \gls{guidedback}, as a consequence of Eq.~\eqref{eq:modified_backprop}, whenever a ReLU is encountered, the backward flow of negative gradients is suppressed, only allowing positive gradients to flow through, thereby focusing on features that positively influence the network activation, as originally proposed in \cite{springenbergStrivingSimplicityAll2015}. 

Notice that, as a result of Eq.~\eqref{eq:guidedb}, a relevance score is obtained for each input feature in the reshaped \gls{ecg} signal, or equivalently, $\mathbf{S}_r,\:\mathbf{B}_c^{\mathbf{S}_r}\in\mathbb{R}^D$ always have the same dimensionality regardless of the \gls{ecg} formatting.

\subsubsection{Grad-CAM}
\gls{gradcam} is a gradient-based explainability method designed to identify the regions of the input that contribute most to a specific class prediction. It does so by establishing a connection between the class-specific logit $z_c$ 
from the final fully connected layer and the feature maps $\mathbf{x}_{\mathcal{L}}^j$, $j = 1,\dots,M_\mathcal{L}$, produced by the last convolutional layer, extracting saliency map $\mathbf{C}_{c}^{\mathbf{S}_r}$ as
\begin{equation}\label{eq:gradcam}
    \mathbf{C}_{c}^{\mathbf{S}_r} = ReLU\bigg(\sum_{j=1}^{M_{\mathcal{L}}} \alpha_{c}^j \cdot \mathbf{x}_{\mathcal{L}}^{j}\bigg),\:\: c=0,\dots, C-1,
\end{equation}
where $\mathbf{x}_{\mathcal{L}}^j$ denotes the $j$-th feature map at the last convolutional block, and the weight $\alpha_{c}^j$ is defined as:
\begin{equation*}
    \alpha_{c}^j = \frac{1}{|\{f\in \mathbf{x}_{\mathcal{L}}^{j}\}|}\sum_{f \in \mathbf{x}_{\mathcal{L}}^{j}} \frac{\partial z_c}{\partial f}.
\end{equation*}
The weights $\alpha_{c}^j$ are obtained by applying \gls{gap} to the gradients of the target logit $z_c$ with respect to each activation $f$ in the feature map $\mathbf{x}_{\mathcal{L}}^{j}$ \cite{selvarajuGradCAMVisualExplanations2020}. This weighting scheme captures the importance of each feature map in predicting class $c$, while the subsequent ReLU nonlinearity filters out negative influences, ensuring the final saliency map highlights only features that positively contribute to the class decision. 

It is important to highlight that the number of features in the feature maps, and therefore the number of scores in the saliency maps, can differ from the total number of input features $L\cdot T$, that is, the number of time step per number of leads. Indeed, as discussed for every \gls{fcn}, the dimensionality of feature maps varied with the \gls{ecg} shaping and with the stride hyperparameter. For this reason, not all of the feature maps were suitable for explaining all the values in the \glspl{ecg}.

Specifically, when using multi-channel \glspl{fcn} with one-dimensional convolutional kernels of Section \ref{subsec:multichannel}, the network extracted one-dimensional feature maps with one feature per time step, i.e. $\mathbf{x}_{\mathcal{L}}^j\in \mathbb{R}^T,\:\forall\:j=1,\dots,M_{\mathcal{L}}$. Their weighted sum resulted in saliency maps that assigned one relevance score per time step, namely $\mathbf{C}_c^{\mathbf{S}_r}\in\mathbb{R}^T$, collapsing the lead dimension and thus preventing separate assessment of each of the 12 leads’ contributions to the classification.

Conversely, in the single-channel \gls{fcn} configuration (Section \ref{subsec:stack}), where all leads were stacked sequentially to form a univariate time series, the feature map dimensionality matched the input dimensionality. However, the use of a stride greater than 1 lead to extracting feature maps shorter than the full length of the stacked input. This discrepancy reduced the number of relevance scores compared to the original input length, necessitating interpolation to restore matching dimensions. Such interpolation risks losing important fine-grained information about the model’s decision function.

Therefore, to obtain saliency maps that are both high-resolution and able to discriminate contributions from each lead at every time step, only the single-channel \gls{fcn} with 2D filters with a stride of 1 provided a suitable framework where $\mathbf{C}_c^{\mathbf{S}_r}\in\mathbb{R}^{T\times L \times 1}$ matched the dimension of the \gls{ecg} signal. This approach preserved both temporal and lead-specific resolution, enabling more interpretable and granular explanations of the model predictions.

\subsubsection{Guided Grad-CAM}
\gls{ggradcam} is a \gls{xai} method that combines the localization capabilities of \gls{gradcam} with the fine-grained sensitivity of \gls{guidedback} to generate high-resolution saliency maps \cite{selvarajuGradCAMVisualExplanations2020}, inheriting the spatial precision of \gls{guidedback} and the class-discriminative properties of \gls{gradcam}. \gls{ggradcam} integrates these two methods by performing an element-wise product between the Guided Backpropagation gradients and the Grad-CAM saliency map, meaning that the final saliency map $\mathbf{M}_c^{\mathbf{S}_r}$ computed for instance $\mathbf{S}_r$ and logit $z_c$ is
\begin{equation*}
    \mathbf{M}_c^{\mathbf{S}_r} = \mathbf{B}_c^{\mathbf{S}_r}\cdot\mathbf{C}_c^{\mathbf{S}_r},\:\: c = 0,\dots,C-1,
\end{equation*}
which can be computed when $\mathbf{B}_c^{\mathbf{S}_r}$ and $\mathbf{C}_c^{\mathbf{S}_r}$ have the same dimensionality, a condition that was met only when the inputs were fed as images.

In this work, we focused on \gls{ggradcam} saliency maps of correctly classified test samples, using the absolute values of the scores. This approach was motivated by the findings of Suh \emph{et al.} \cite{suhVisualInterpretationDeep2024} in the context of pathological \gls{ecg} classification, where the regions highlighted by the absolute \gls{ggradcam} scores were shown to well correspond with features identified by medical experts as clinically relevant. 
As a preliminary step to interpreting the network's decisions, the \glspl{ecg} within a given region were partitioned into clusters as described in Section \ref{subsec:clustering}, in order to refine the analysis to account for morphological differences in same-class data, possibly affecting highlighted \gls{ecg} segments in the saliency maps.    
Subsequently to clustering, the saliency maps were visually inspected by the expert in \gls{ecg} generation through computational methods responsible of the data generation, to understand whether the features driving the network's choices could be linked to a physiological interpretation. Further considerations were made to assess the effectiveness of the \gls{guidedback} and \gls{gradcam} methods in highlighting pathology-related patterns.

\subsubsection{Physiological Interpretation of XAI Results}\label{subsec:clustering}
To assess the relationship between \gls{ecg} morphological features highlighted by the \gls{xai} methods and the underlying \gls{ep} activity, a qualitative visual analysis was conducted. This analysis aimed to correlate the evolution of the transmembrane voltage signal in the virtual heart model with the corresponding \gls{ecg} segments identified by the \gls{xai} techniques. Since the 24 anatomical regions were defined \emph{a priori}, substantial variability in \gls{ecg} morphology was observed among signals associated with the same region, making it challenging to establish a generalized correlation between highlighted \gls{ecg} features and specific anatomical locations. These variations were also reflected in the \gls{xai} outputs. To address this limitation, we applied clustering within each region to group \glspl{ecg} with similar morphological patterns, thereby enabling a more consistent and interpretable mapping between \gls{ecg} features and \gls{ap} locations.

We employed the K-medoids clustering algorithm in combination with the \gls{dtw} distance \cite{muller2007dynamic} to group signals with similar shapes, regardless of temporal shifts. This approach allowed us to cluster \glspl{ecg} based exclusively on their morphological features, consistently with the focus of the \gls{xai} techniques, which highlight relevant regions of the waveform without considering absolute timing. 
K-medoids selects a real \gls{ecg} signal as the cluster representative, preserving physiological interpretability while also reducing the computational cost associated with centroid estimation \cite{izakian2015fuzzy}. For each anatomical region, we explored a range of cluster numbers $k \in \{2,3,4\}$, and selected the optimal $k$ by maximizing the average silhouette score \cite{belyadi2021unsupervised}, which quantifies both intra-cluster similarity and inter-cluster separation.

\subsubsection{Lead importance}
The \gls{xai} method used in this work produces a relevance score for each time step and each lead of the input \gls{ecg}, resulting in a normalized importance map across all leads. While originally intended to explain individual classifications, this map can also be exploited to uncover global morphological patterns and guide the development of future diagnostic strategies. Specifically, it enables the identification of the most influential leads, those whose variations most strongly contribute to the model predictions. To quantify this, we define an \emph{index of lead importance} that summarizes the information extracted from the \gls{xai} maps across the test set.

Intuitively, for a sample correctly classified as belonging to class $c$, the higher the relevance scores associated with a given lead, the greater its contribution to the classification decision. To evaluate this systematically across all correctly classified samples, we define the lead importance for each lead $l = 1,\dots,L$ and class $c = 0,\dots, C-1$ as:

\begin{equation}\label{eq:lead_importance}
    LI_l^c = \frac{\sum_{\mathbf{S}_r \in C_c}\sum_{t=1}^T (M_c^{\mathbf{S}_r})_{tl}}{\sum_{l=1}^L\sum_{\mathbf{S}_r \in C_c}\sum_{t=1}^T (M_c^{\mathbf{S}_r})_{tl}},
\end{equation}

where $(M_c^{\mathbf{S}_r})_{tl}$ denotes the absolute value of the \gls{ggradcam} relevance score at time step $t$ in lead $l$, for the sample $\mathbf{S}_r$ belonging to the set $C_c$ of test samples correctly assigned to class $c$. The results of this analysis are presented in Section~\ref{subsec:explainability_results}.


\section{Results}
\label{Sec:results}
All of the \gls{dl} models described in Sections \ref{subsec:stack}-\ref{subsec:image} were first tuned through Bayesian Optimization, then, using the optimal values of hyperparameters fixed with tuning and reported in \ref{sec:hyperparameters}, models were trained and fine-tuned at the last layers to further refine the higher-level extracted representations. The \gls{ecg} dataset was split into training, validation, and test sets with a proportion of (75, 15, 10)\%. Selected hyperparameters for tuning were the kernel size, the number of convolutional filters, and, in the single case of Section \ref{subsec:stack}, the stride. The number of hidden layers was prescribed to be 3, as for the \gls{fcn} in \cite{fawazDeepLearningTime2019}.

\subsection{Classification performance}\label{subsec:classification}
Following tuning and fine-tuning, both the multi-channel \gls{fcn} and the single-channel \gls{fcn} with 2D kernels achieved comparable accuracies on the test set -- 95.48\% and 95.03\%, respectively. In contrast, treating \glspl{ecg} as a stack of leads resulted in a lower performance of 91.98\%. Sensitivity and specificity were computed for the \gls{fcn} with best accuracy performances, and listed in Table~\ref{tab:class_metrics} for the interpreted 2D \gls{fcn} model. Excellent classification is achieved for most regions, including most of the septal ones. 

Reduced sensitivity was displayed in regions 2, 9, 10 and 23, with the latter three corresponding to the mid and mid-ventricular segments in the bi-ventricular mesh, as displayed in Figures \ref{fig:lv}-\ref{fig:rv}. In these cases, false negatives were frequently classified into the anatomically corresponding basal region within the same rotational coordinate interval. This misclassification pattern is attributable to the lower density of sampled points in the mid-ventricular region, which resulted from physiological constraints imposed by the computational model used for data generation, in conjunction with class stratification being preserved during the data split.

\begin{figure}
    \centering
    \includegraphics[width=\linewidth]{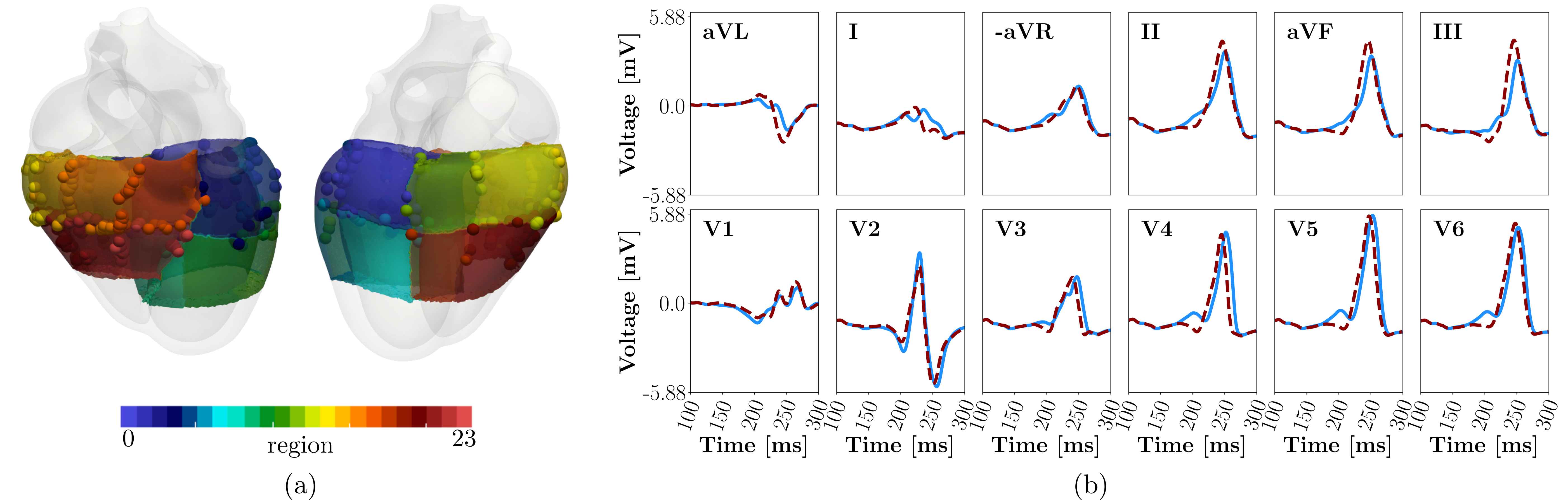}
    \caption{Data misclassified by \gls{fcn} in Section \ref{subsec:image}. (a) \glspl{ap} of misclassified data are plotted with the color of the region they have been assigned to. (b) Comparison of a misclassified test sample (dashed) with a correctly classified training sample (continuous) assigned to the same class.}
    \label{fig:misclassifications}
\end{figure}

Visual inspection of the data misclassified by the single-channel \gls{fcn} with 2D kernels revealed that most errors occurred, understandably, for \glspl{ecg} corresponding to \glspl{ap} located near the boundary between two or more adjacent regions, as illustrated in Figure~\ref{fig:misclassifications}a. Notably, out of 44 misclassified test samples, 39 had the second largest predicted probability corresponding to the true class.

A specific analysis was performed to evaluate the ability of the 2D \gls{fcn} to identify the septal antegrade region, located near the His bundle and the \gls{av} node, which are traditionally difficult to localize \cite{arrudaDevelopmentValidationECG1998, elhamritiEASYWPWNovelECGalgorithm2023}. To this end, regions 3, 5, 9, and 11 were grouped together due to their proximity to the His bundle, while regions 1, 2, 7, and 8 were grouped to encompass the medial-posterior septum. The first set of regions achieved a sensitivity of 92.41\% and a specificity of 99.04\%, whereas the latter set achieved 92.86\% sensitivity and 98.93\% specificity.


\begin{table}[t!]
\centering
\begin{tabular}{c|c|c|c|c}
\textbf{Class} & \textbf{Specificity (\%)} & \textbf{Sensitivity (\%)} & \textbf{Train cardinality} & \textbf{Test cardinality} \\
\midrule
0  & 99.51 & 93.44 & 457 & 61 \\
1  & 99.27 & 100.00 & 457 & 61  \\
\cellcolor{LightYellow}2  & \cellcolor{LightYellow}99.76 & \cellcolor{LightYellow}88.37 & \cellcolor{LightYellow}316 & \cellcolor{LightYellow}43  \\
3  & 99.51 & 92.42 & 490 & 66  \\
4  & 99.28 & 98.18 & 415 & 55  \\
5  & 99.88 & 92.98 & 424 & 57  \\
6  & 99.88 &92.86 & 213 & 28  \\
7  & 100.00 & 90.00 & 146 & 20  \\
8  & 99.66 & 92.86 & 110 & 14  \\
\cellcolor{LightYellow}9  & \cellcolor{LightYellow}99.88 & \cellcolor{LightYellow}83.33 & \cellcolor{LightYellow}140 & \cellcolor{LightYellow}18  \\
\cellcolor{LightYellow}10 & \cellcolor{LightYellow}100.00 & \cellcolor{LightYellow}81.82 & \cellcolor{LightYellow}84  & \cellcolor{LightYellow}11  \\
11 & 99.88 & 100.00 & 126 & 17  \\
12 & 100.00 & 93.88 & 367 & 49  \\
13 & 99.76 & 96.30 & 402 & 54  \\
14 & 99.76 & 100.00 & 322 & 43  \\
15 & 99.88 & 100.00 & 327 & 44  \\
16 & 99.88 & 96.55 & 434 & 58  \\
17 & 99.52 & 97.83 & 344 & 46  \\
18 & 100.00 & 100.00 & 113 & 15  \\
19 & 99.88 & 96.97 & 251 & 33  \\
20 & 100.00 & 95.45 & 164 & 22  \\
21 & 99.88 & 100.00 & 149 & 20  \\
22 & 99.88 & 95.83 & 184 & 24  \\
\cellcolor{LightYellow}23 & \cellcolor{LightYellow}99.77 & \cellcolor{LightYellow}84.62 & \cellcolor{LightYellow}196 & \cellcolor{LightYellow}26  \\
\end{tabular}
\caption{Class-wise specificity, sensitivity, and cardinality.}
\label{tab:class_metrics}
\end{table}

\subsubsection{Comparison with the EASY-WPW and Arruda algorithms}
We compare the performance of the EASY-WPW and Arruda \gls{dt} algorithms with our \gls{fcn}
by regrouping the \glspl{ecg} according to the anatomical regions defined by each \gls{dt} method, and we manually classify the \glspl{ecg} following the corresponding \gls{dt} rules. A one-sided Fisher’s exact test was finally performed. On the EASY-WPW regions, the \gls{fcn} correctly classified 100\% of the subset samples, whereas the \gls{dt} achieved only 76\% accuracy (p-value = $1.20 \cdot 10^{-6}$). Similarly, the network demonstrated superior performance on the Arruda regions, attaining 100\% accuracy compared to 72\% for the Arruda algorithm (p-value = $9.36 \cdot 10^{-8}$).


\subsection{Physiological Interpretation of XAI Results}\label{subsec:explainability_results}
\begin{figure}[htbp]
    \centering
    \includegraphics[width=0.82\textwidth]{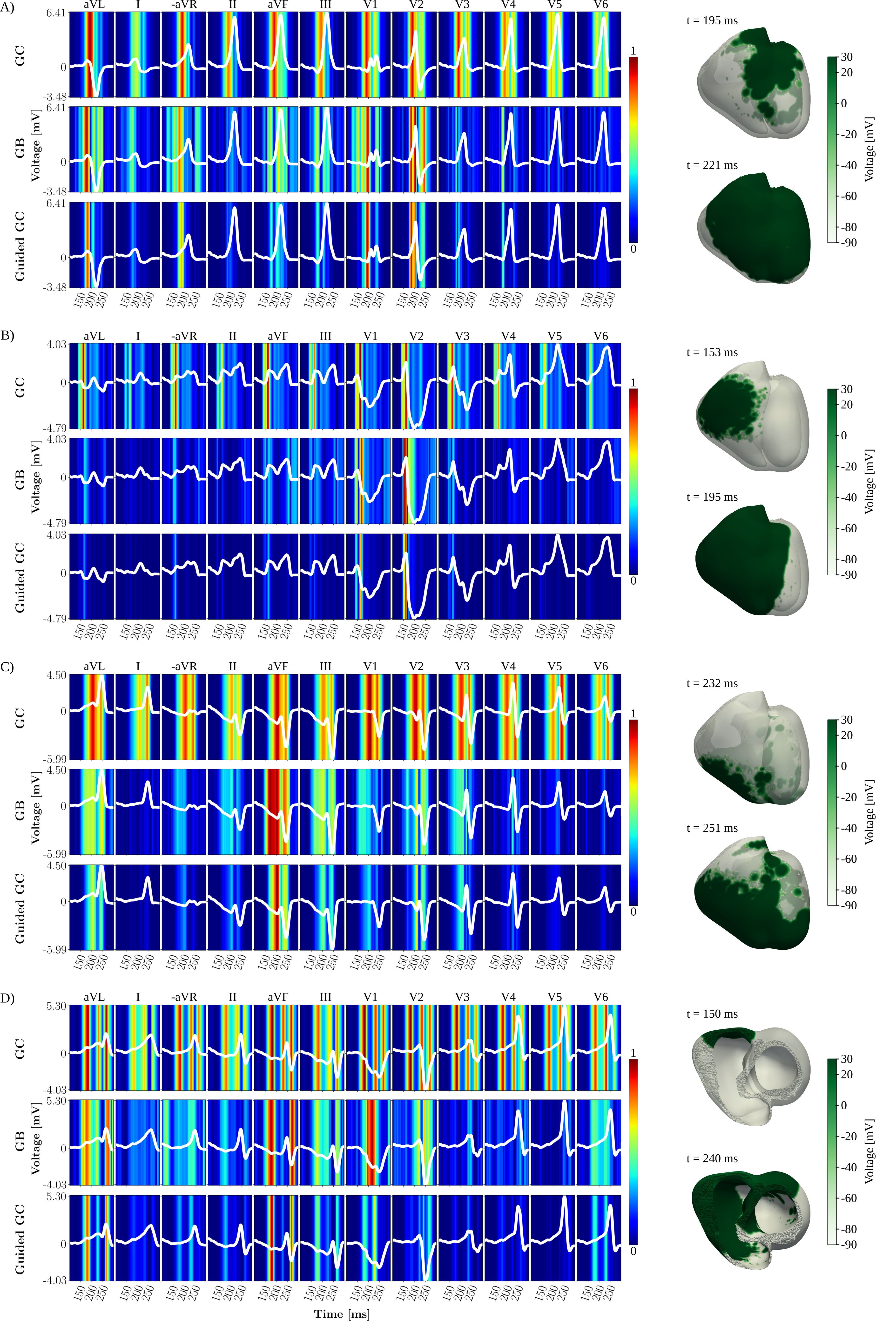}
    \caption{\gls{gradcam} (GC), \gls{guidedback} (GB), and \gls{ggradcam} (Guided GB) saliency maps, and activation patterns across ECG leads vary with the anatomical location and conduction behavior of \glspl{ap}.
(A) \gls{ap} contributing to the ongoing wavefront highlights features in the QRS complex in multiple leads, especially aVL and aVF.
(B) Right ventricular \gls{ap} demonstrates localized activation in leads V1 and V2, consistent with activation disruptions from the RV free wall.
(C) \gls{ap} initiating early ventricular activation produces diffuse explainability across several leads.
(D) \gls{ap} in the RV free wall outside the His-Purkinje system reveals two distinct ECG regions: the delta wave and His-Purkinje capture.}
    \label{fig:XAi_explainability}
\end{figure}

\begin{figure}[t!]
    \centering
    \includegraphics[width=0.9\linewidth]{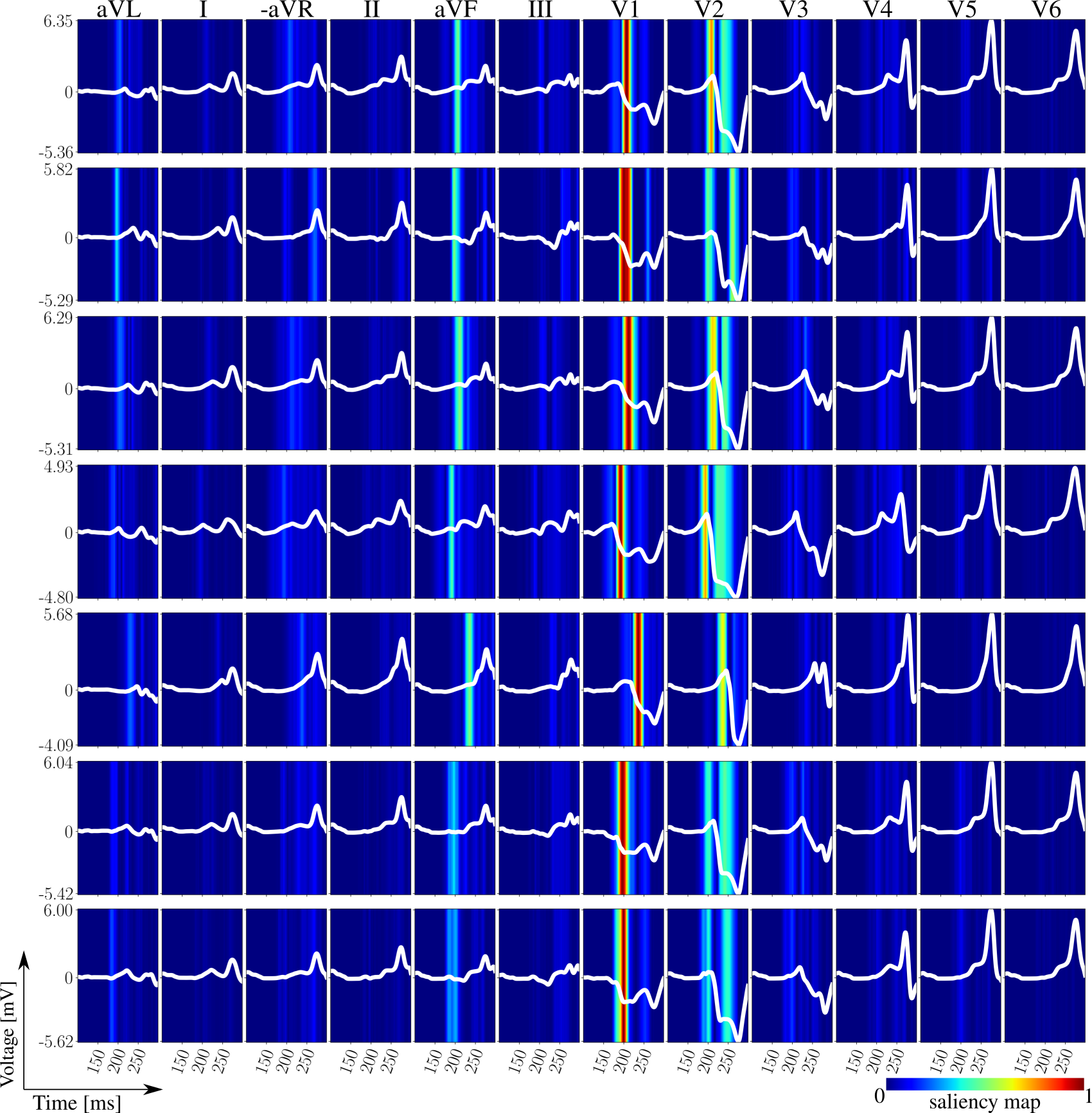}
    \caption{Example of \gls{ecg} variation for a cluster in the right ventricle (region 1), and corresponding saliency maps computed with the \gls{ggradcam}.}
    \label{fig:xai_rv}
\end{figure}

\begin{figure}[t!]
    \centering
    \includegraphics[width=0.9\linewidth]{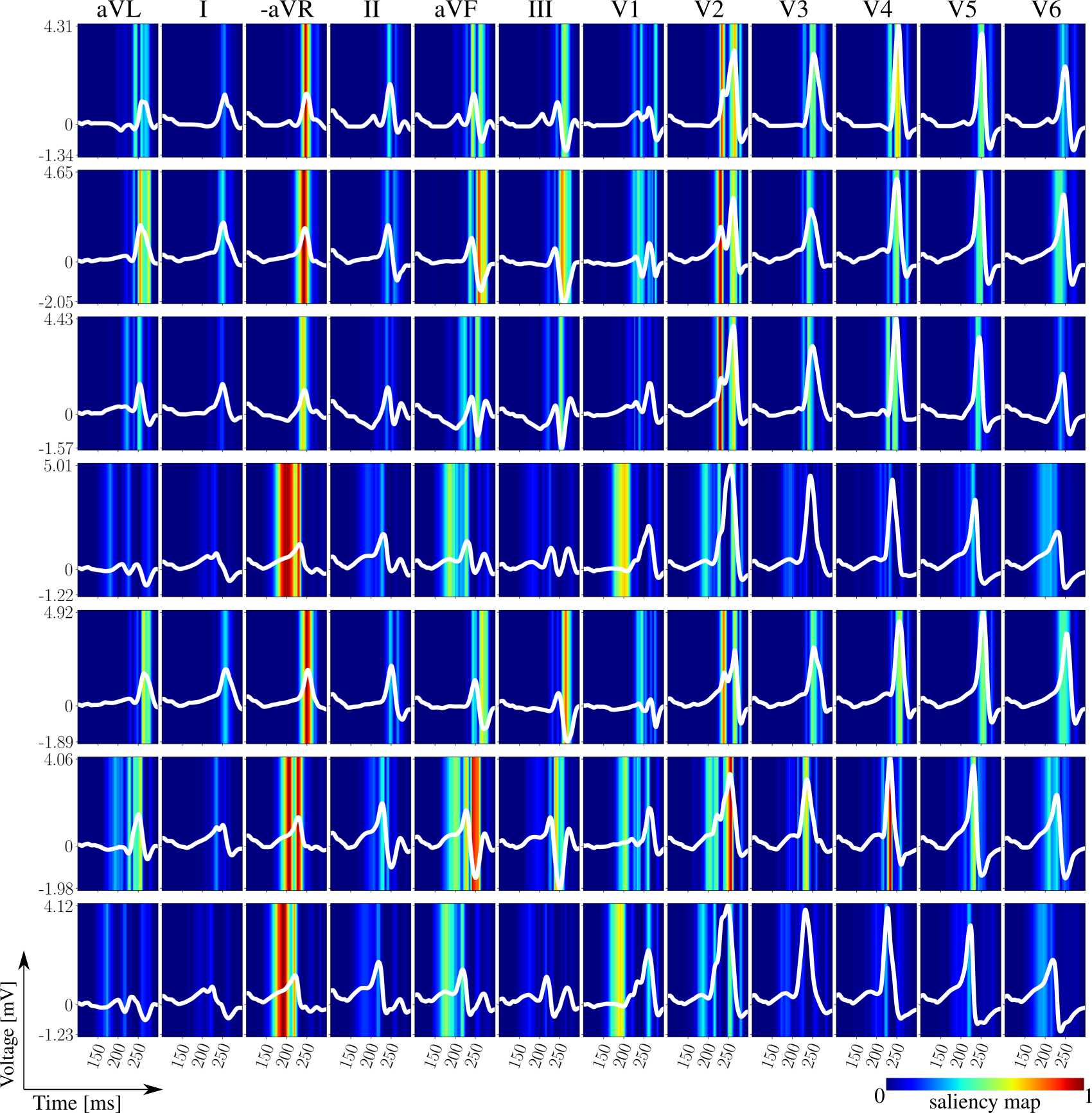}
    \caption{Example of \gls{ecg} variation for a cluster in the left ventricle (region 1), and corresponding saliency maps computed with the \gls{ggradcam}.}
    \label{fig:xai_lv}
\end{figure}

Saliency maps derived from all three evaluated \gls{xai} techniques revealed consistent identification of morphologically relevant features across the 12-lead \gls{ecg} that serve as biomarkers in \gls{wpw} syndrome, corresponding to the anatomical location of the \gls{ap} across various cardiac regions, though the extent and localization of activations varied across clusters (Figure~\ref{fig:XAi_explainability}). Within the saliency representations, \glspl{ecg} associated with right ventricular \glspl{ap} formed distinct clusters characterized by stable morphological patterns, whereas left ventricular \glspl{ap} exhibited greater heterogeneity (we referred to examples of saliency maps with \gls{ggradcam} for the right and left ventricles in  Figures~\ref{fig:xai_rv} and~\ref{fig:xai_lv}). Despite inter-cluster variability, all \gls{xai} techniques consistently highlighted key morphological features - most notably the QRS complex and delta wave - across all cases, underscoring their diagnostic relevance in identifying \gls{ap}-mediated conduction abnormalities.

More specifically, across all clusters, \gls{gradcam} produced broad activations spanning multiple leads - particularly within the QRS complex - reflecting the generally altered ventricular activation patterns induced by the \gls{ap} (Figure~\ref{fig:XAi_explainability}). In contrast, \gls{guidedback} demonstrated greater spatial selectivity, emphasizing specific leads while still targeting similar temporal regions of the signal. The \gls{ggradcam} approach, which combines the coarse localization of \gls{gradcam} with the fine-grained sensitivity of guided backpropagation, yielded the most precise identification of diagnostically relevant features. This method produced highly localized activations that aligned closely with the underlying \gls{ap} morphology and its electrophysiological impact on the 12-lead \gls{ecg}.

Further physiological analysis and interpretation revealed distinct patterns of \gls{xai} activation depending on the anatomical location, timing,
and the conduction behavior of the \gls{ap}. When the AP was positioned and timed such that it contributed to the ongoing wavefront of cardiac propagation (Figure \ref{fig:XAi_explainability}A), \gls{xai} methods reliably highlighted corresponding features in the onset of the QRS complex across multiple leads, most notably aVL and aVF.
In right ventricular \glspl{ap}, localized activations appeared predominantly in leads V1 and V2 (Figure~\ref{fig:XAi_explainability}B), consistent with expected disruptions originating from the right ventricular free wall. Conversely, when \glspl{ap} initiated premature ventricular contraction from a site not integrated into the normal conduction pathway, the \gls{xai} techniques highlighted extended segments across several leads (Figure \ref{fig:XAi_explainability}C), indicating a more diffuse impact on cardiac morphology. This is further shown with an \gls{ap} located in the \gls{rv} free wall outside of the His-Purkinje system (Figure \ref{fig:XAi_explainability}D). The XAI techniques detect two primary portions at different times of the ECG, corresponding to the delta wave and directly during His-Purkinje capture. 
Collectively, these patterns highlight how \gls{xai}-derived saliency maps capture both spatial and temporal characteristics of \gls{ap}-mediated conduction, reflecting underlying electrophysiological mechanisms observed in \gls{wpw} syndrome.

\subsection{Lead importance}
\label{subsec:important_leads}
\begin{figure}[!t]
    \centering
    \includegraphics[height=6.5cm, keepaspectratio]{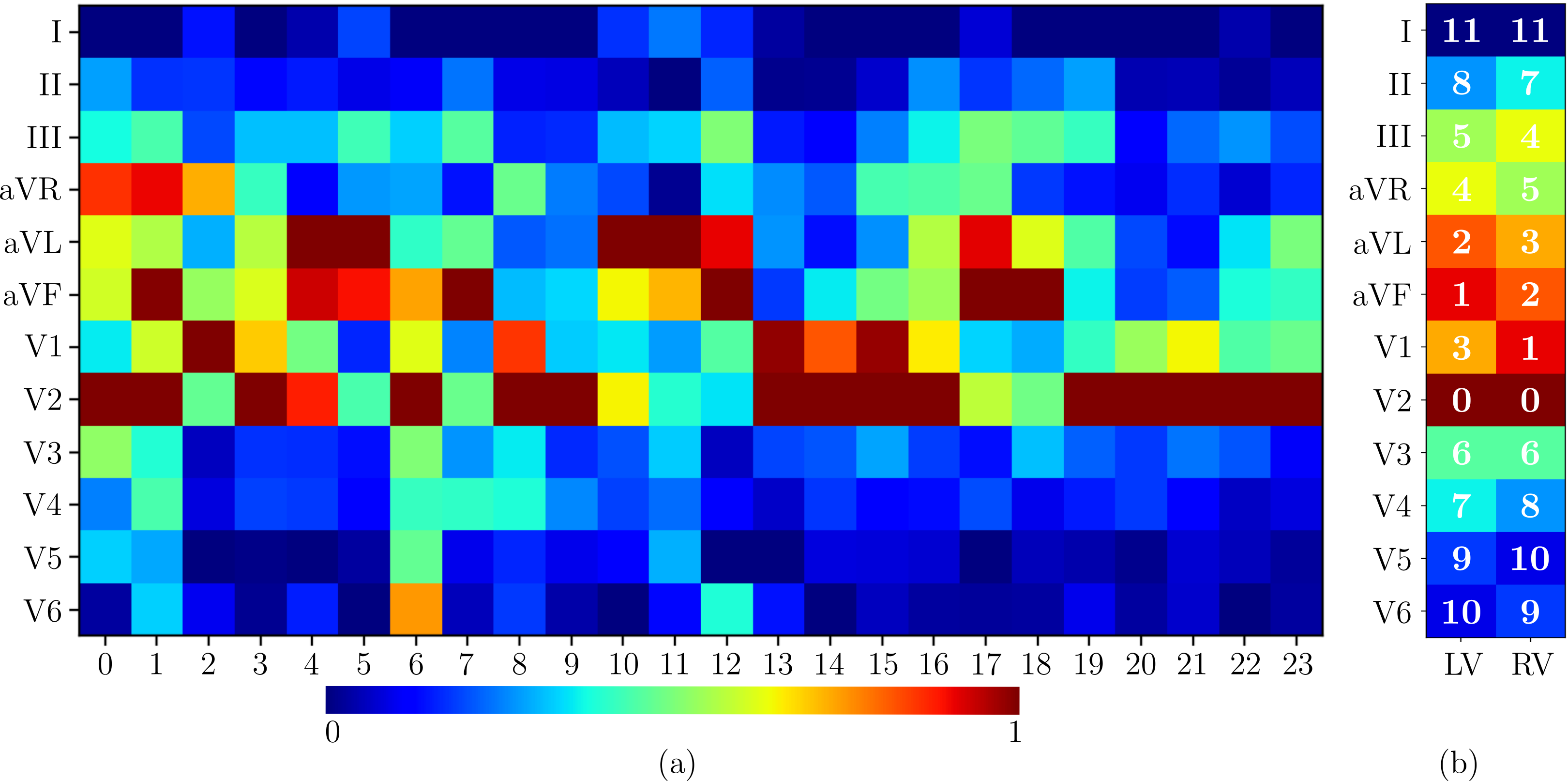}
    \caption{Lead importances according to Eq.\eqref{eq:lead_importance}. (a) Lead importance for each of the 24 regions. (b) Lead importance grouping \gls{lv} and \gls{rv} signals, ranked from most (0) to least (11) important.}
    \label{fig:lead_importance}
\end{figure}

Following the \gls{xai} results and interpretation, we focused on investigating the \gls{ggradcam} method to further quantify which \gls{ecg} leads most frequently contained features deemed influential by the \gls{xai} method in the model’s class predictions. 
The lead importance $LI_l^c$ in ~\eqref{eq:lead_importance} is reported for each lead-class pair as a heatmap in Figure~\ref{fig:lead_importance}a.
The distribution of lead contributions was non-uniform across regions, indicating that specific leads play a more dominant role in guiding classification. Among these, leads V2, aVF, aVL, and V1 consistently exhibited the highest importance across multiple anatomical classes. 

To enable regional comparison, signals were further grouped by ventricular origin of the \gls{ap}, and aggregated lead importance indices were computed for left and right ventricular \glspl{ap} and displayed in Figure~\ref{fig:lead_importance}b). These ventricular-level summaries highlight consistent intraventricular patterns of lead relevance, with V2 remaining the most important lead for both right and left ventricular \gls{ap}, followed by leads V1, aVL, and aVF, albeit in a different order of importance depending on the considered ventricle.



\section{Discussion}\label{sec:discussion}

\subsection{Classification performance}\label{subsec:classification_discussion}

In this work, we investigate a highly effective deep learning–based model for the localization of \glspl{ap}. Among the three evaluated architectures, two achieved accuracies exceeding 95\%, outperforming most previously reported \gls{ap} localization algorithms, including both \gls{dt}- and \gls{ml}-based approaches~\cite{arrudaDevelopmentValidationECG1998,elhamritiEASYWPWNovelECGalgorithm2023,pambrunMaximalPreExcitationBased2018,senonerIdentifyingLocationAccessory2021,nishimoriAccessoryPathwayAnalysis2021,yahyazadehNovelFeatureExtraction2024}. The performance improvements over existing \gls{dt}-based algorithms were further validated by comparing the \gls{fcn} model with manual classifications of our \glspl{ecg} using two representative \glspl{dt}.
The \gls{fcn} achieved 100\% accuracy, markedly exceeding the performance of the two \gls{dt}-based methods, which reached 76\% and 72\% overall accuracy.

Although the \gls{dt}-based model by Khalaph \emph{et al.}~\cite{khalaph2025novel} achieved a marginally higher accuracy of 97\%, it was restricted to 12 manually defined regions describing only the rotational position of the \gls{ap}. In contrast, our approach considers 24 classes, introducing a longitudinal subdivision that enhances the spatial resolution of \gls{ap} localization. Furthermore, the classification regions were defined using standardized \gls{uvc} coordinates, ensuring reproducibility and anatomical consistency across subjects.

Three \gls{cnn} architectures sharing the same core structure but differing in input dimensionality were implemented and systematically evaluated to optimize localization performance. To mitigate data scarcity and enhance training variability, a virtual dataset was generated using a cardiac digital twin, simulating \glspl{ecg} corresponding to \glspl{ap} at multiple cardiac locations. This strategy effectively overcomes a key limitation of previous studies~\cite{senonerIdentifyingLocationAccessory2021,middlehurst2024bake}, where the small number of available \glspl{ecg} constrained both the training set size and the number of anatomical regions that could be defined as classes. By generating synthetic \glspl{ecg} through the digital twin, our approach expands the dataset while improving its diversity, thereby enhancing the generalization capability of the \gls{fcn} to unseen data.

To the best of our knowledge, this study provides the first rigorous analysis of how different input data structures affect \gls{ecg}-based classification, explicitly considering lead differentiation and the identification of lead-specific importance during model inference. Our results demonstrate that optimal performance with the \gls{fcn} is achieved when both single-lead information and temporal dependencies are explicitly modeled. This configuration is realized in the multi-channel \gls{fcn}, where the feature maps capture lead-wise and temporal relationships for each lead, and in the image-like 2D \gls{fcn}, where square kernels jointly encode lead (width) and temporal (height) dependencies. In contrast, the classifier based on a single stacked input of leads exhibited the lowest accuracy, reflecting the absence of explicit spatial and temporal characterization. Future improvements could involve the use of variable-length or dilated kernels across the lead stack, as proposed by Ayano \emph{et al}.~\cite{ayano2024interpretable}.


\subsection{Explainability}\label{subsec:explainability_discussion}

To ensure model transparency and physiological interpretability, three \gls{xai} techniques were implemented and rigorously evaluated. The resulting saliency maps demonstrated a strong correspondence between the model’s highlighted features and established electrophysiological patterns, confirming that the network localized \glspl{ap} based on physiologically meaningful signal propagation across the 12-lead \gls{ecg} (Figures~\ref{fig:XAi_explainability},~\ref{fig:xai_rv},~\ref{fig:xai_lv}). These findings validate the internal consistency of the \gls{fcn} and its ability to leverage morphological information in a manner coherent with known cardiac conduction mechanisms.

Model interpretability can be assessed at two complementary levels: individual predictions and aggregate population trends. Most existing explainable \gls{ecg} classification studies~\cite{vandeleurDiscoveringVisualizingDiseaseSpecific2021, leLightX3ECGLightweightEXplainable2022, jones2020improving, xie2024intelligent} have focused primarily on the former, visually inspecting the salient regions for single cases or diagnostic classes, often without quantifying broader lead-level trends across datasets. In contrast, the present study integrates both perspectives. Following methodologies similar to Ayano \emph{et al.}~\cite{ayano2024interpretable} and Agrawal \emph{et al.}~\cite{agrawal2022ecg}, saliency maps were first analyzed qualitatively, then summarized quantitatively through the computation of a lead-importance index for each lead–class pair. This metric provides a reproducible, dataset-level measure of the relative influence of each lead in model decision-making, a quantity not formalized in previous works, but only qualitatively observed.

To ensure accurate and interpretable saliency mapping, the 2D-kernel \gls{fcn} was used, as it guarantees that saliency maps match the input signal dimensions. Previous methods~\cite{leLightX3ECGLightweightEXplainable2022, ayano2024interpretable} processed each lead independently using 1D CNN backbones, combining the resulting feature maps through attention mechanisms or concatenation. While effective, this joint optimization introduces dependencies across leads, meaning that the relevance scores for a given lead cannot be considered fully independent. By contrast, processing the ECG as an image with a 2D \gls{fcn} provides a straightforward and physiologically consistent framework for saliency analysis across all 12 leads, avoiding the added complexity of multi-backbone architectures.

Visual inspection of the saliency maps revealed that activation patterns varied systematically with \gls{ap} location and conduction characteristics. For right ventricular pathways, saliency concentrated around precordial leads V1 and V2, while left-sided pathways were characterized by stronger activity in aVL and aVF. These trends are consistent with the spatial projections of ventricular depolarization on the standard 12-lead \gls{ecg}. Quantitative analysis of lead importance further confirmed this pattern: leads V2, aVF, aVL, and V1 consistently emerged as the most influential across multiple anatomical classes. At the ventricular level, V2 remained the most important lead for both right and left \glspl{ap}, followed by V1, aVL, and aVF, albeit in varying order of relevance depending on the ventricle.

The pronounced importance of the augmented Goldberger leads (aVL, aVF, aVR) compared with the standard limb leads (I, II, III) likely stems from their unipolar configuration. Unlike bipolar limb leads, which measure the potential difference between two electrodes, Goldberger leads reference the Wilson central terminal, a composite average of the other limb electrodes. This effectively amplifies signal magnitude and enhances sensitivity to subtle morphological variations, which deep learning models can exploit more efficiently. Consequently, the network may capture electrophysiological nuances in these leads that are less discernible in standard limb leads.

Importantly, the ranking of influential leads derived from the \gls{xai} analysis aligns closely with independent sensitivity studies. Gillette \emph{et al.}~\cite{gilletteComputationalStudyInfluence2025} similarly identified V2 as the most sensitive lead for accessory pathway localization in the same dataset. This convergence supports the physiological validity of the explainability framework and suggests that the \gls{fcn} internal representations reflect true cardiac conduction dynamics rather than spurious correlations. Nonetheless, this observation should be interpreted with caution: the prominence of V2 may be partially influenced by the specific heart orientation and electrode positioning in the subject used to generate the data. Expanding the dataset to include a broader range of anatomical and geometric configurations could provide further insight into the generalizability of these findings.

Finally, by identifying a subset of leads with high diagnostic relevance, the lead-importance framework provides practical insights for future model design. Training lighter \gls{dl} architectures on low dimensional input centered on these key leads could preserve diagnostic accuracy while improving computational efficiency. More broadly, the proposed approach illustrates how explainability metrics can bridge data-driven inference and physiological understanding, ensuring that model predictions remain interpretable and clinically grounded.

\section{Limitations}
\label{sec:limitaions}
The main limitation of this work lies in the dataset employed to train the \gls{fcn}. While rich in terms of the number of \gls{ecg} traces and possible variability depending on the \gls{ap} position, it was restricted to a single subject, thus strongly limiting possible natural inter-patient variability. This hinders the generalization ability of the network to unseen data coming from cardiac digital twins of other patients and clinical data. The scope of future work will be to augment this dataset, including more subjects, and to test the obtained network on real clinical data.



The dataset used for training and evaluation was generated from a single subject’s anatomy, which constrains the anatomical variability captured in the study and may influence lead-specific importance patterns, such as the prominent role of V2. Although the virtual dataset created via a cardiac digital twin increases the diversity of training examples, it cannot fully reproduce the anatomical and electrophysiological heterogeneity present in a broader population. Expanding the model to real patient data will be essential to validate generalizability and confirm the physiological relevance of observed lead-importance trends.

Further limitations arise from the limited theoretical foundation of post-hoc explainability methods in the context of time series data. While interpreting saliency maps for image data is relatively straightforward due to the explicit semantics (e.g., edges, colors), physiological time series such as \glspl{ecg} are not yet fully understood, and the relationship between the \gls{ecg} and the underlying electrophysiology remains unclear.
Moreover, even with 2D-kernel \glspl{fcn}, the relevance scores for individual leads may be influenced by interactions with other leads, limiting the independence of lead-specific interpretations. No formal mathematical framework currently exists to validate the physiological significance of saliency map features, making expert clinical evaluation necessary. These limitations will be addressed in future work to strengthen both the robustness and interpretability of the model.

\section{Conclusions}
\label{sec:conclusions}
In this work, we present the first \gls{dl}-based method, implemented as a \gls{fcn}, for the localization of antegrade \glspl{ap} within the ventricles. Our approach outperforms existing \gls{dt}- and \gls{ml}-based techniques in both accuracy and spatial precision, enabled by a higher number of anatomical classes. \gls{xai} techniques were applied to enhance model transparency, revealing a clear correspondence between predictions and underlying electrophysiology. A new index was introduced to identify the most influential lead in \gls{ap} classification, exposing potential limitations in current localization strategies. These results demonstrate that explainable deep learning can provide highly accurate, physiologically grounded \gls{ap} localization, offering a promising tool to support clinical decision-making and improve diagnostic workflows.

\section*{Acknowledgments}
This research received support from the Austrian Science Fund (FWF) grant no. 10.55776/I6476, and from the European Union’s Horizon 2020 research and innovation program under the Marie Sk\l{}odowska-Curie grant TwinCare-AF (grant agreement no. 101148636).
\begin{figure}[!h]
    \centering
    \includegraphics[width=0.5\linewidth]{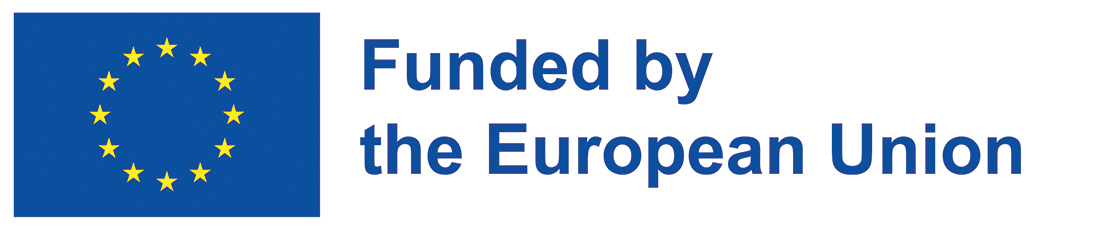}
    \label{fig:placeholder}
\end{figure}
E. Zappon and S. Fresca acknowledge her membership to INdAM GNCS - Gruppo Nazionale per il Calcolo Scientifico (National Group for Scientific Computing, Italy). S. Fresca acknowledges the project FAIR (Future Artificial Intelligence Research), funded by the NextGenerationEU program within the PNRR-PE-AI scheme (M4C2, Investment 1.3, Line on Artificial Intelligence).

\bibliographystyle{elsarticle-num}
\bibliography{Ref}

@article{lindsay1987concordance,
  title={Concordance of distinguishing electrocardiographic features during sinus rhythm with the location of accessory pathways in the Wolff-Parkinson-White syndrome},
  author={Lindsay, Bruce D and Crossen, Karl J and Cain, Michael E},
  journal={The American journal of cardiology},
  volume={59},
  number={12},
  pages={1093--1102},
  year={1987},
  publisher={Elsevier}
}

@article{fitzpatrick1994new,
  title={New algorithm for the localization of accessory atrioventricular connections using a baseline electrocardiogram},
  author={Fitzpatrick, Adam P and Gonzales, Rolando P and Lesh, Michael D and Lee, Randall J and Scheinman, Melvin M and others},
  journal={Journal of the American College of Cardiology},
  volume={23},
  number={1},
  pages={107--116},
  year={1994},
  publisher={Elsevier}
}

@article{qian2025developing,
  title={Developing cardiac digital twin populations powered by machine learning provides electrophysiological insights in conduction and repolarization},
  author={Qian, Shuang and Ugurlu, Devran and Fairweather, Elliot and Toso, Laura Dal and Deng, Yu and Strocchi, Marina and Cicci, Ludovica and Jones, Richard E and Zaidi, Hassan and Prasad, Sanjay and others},
  journal={Nature Cardiovascular Research},
  volume={4},
  number={5},
  pages={624--636},
  year={2025},
  publisher={Nature Publishing Group UK London}
}

@article{bhagirath2024bits,
  title={From bits to bedside: entering the age of digital twins in cardiac electrophysiology},
  author={Bhagirath, Pranav and Strocchi, Marina and Bishop, Martin J and Boyle, Patrick M and Plank, Gernot},
  journal={Europace},
  volume={26},
  number={12},
  pages={euae295},
  year={2024},
  publisher={Oxford University Press UK}
}

@incollection{loewe2022cardiac,
  title={Cardiac digital twin modeling},
  author={Loewe, Axel and Mart{\'\i}nez D{\'\i}az, Patricia and Nagel, Claudia and S{\'a}nchez, Jorge},
  booktitle={Innovative treatment strategies for clinical electrophysiology},
  pages={111--134},
  year={2022},
  publisher={Springer}
}

@article{doste2022training,
  title={Training machine learning models with synthetic data improves the prediction of ventricular origin in outflow tract ventricular arrhythmias},
  author={Doste, Ruben and Lozano, Miguel and Jimenez-Perez, Guillermo and Mont, Lluis and Berruezo, Antonio and Penela, Diego and Camara, Oscar and Sebastian, Rafael},
  journal={Frontiers in Physiology},
  volume={13},
  pages={909372},
  year={2022},
  publisher={Frontiers Media SA}
}

@article{gillette2023medalcare,
  title={MedalCare-XL: 16,900 healthy and pathological synthetic 12 lead ECGs from electrophysiological simulations},
  author={Gillette, Karli and Gsell, Matthias AF and Nagel, Claudia and Bender, Jule and Winkler, Benjamin and Williams, Steven E and B{\"a}r, Markus and Sch{\"a}ffter, Tobias and D{\"o}ssel, Olaf and Plank, Gernot and others},
  journal={Scientific Data},
  volume={10},
  number={1},
  pages={531},
  year={2023},
  publisher={Nature Publishing Group UK London}
}

@article{dasi2024silico,
  title={In Silico TRials guide optimal stratification of ATrIal FIbrillation patients to Catheter Ablation and pharmacological medicaTION: the i-STRATIFICATION study},
  author={Das{\'\i}, Albert and Nagel, Claudia and Pope, Michael TB and Wijesurendra, Rohan S and Betts, Timothy R and Sachetto, Rafael and Loewe, Axel and Bueno-Orovio, Alfonso and Rodriguez, Blanca},
  journal={Europace},
  volume={26},
  number={6},
  pages={euae150},
  year={2024},
  publisher={Oxford University Press UK}
}

@article{corral2020digital,
  title={The ‘Digital Twin’to enable the vision of precision cardiology},
  author={Corral-Acero, Jorge and Margara, Francesca and Marciniak, Maciej and Rodero, Cristobal and Loncaric, Filip and Feng, Yingjing and Gilbert, Andrew and Fernandes, Joao F and Bukhari, Hassaan A and Wajdan, Ali and others},
  journal={European heart journal},
  volume={41},
  number={48},
  pages={4556--4564},
  year={2020},
  publisher={Oxford University Press}
}

@article{de2023machine,
  title={Machine learning in cardiac electrophysiology},
  author={de Oliveira Bernardo, Diana Sofia},
  year={2023}
}

@article{hennecken2025localization,
  title={Localization of accessory pathways in Wolff-Parkinson-white syndrome using ECG-based multi-task deep learning},
  author={Hennecken, Jasper and Arends, Bauke KO and Mast, Thomas and Dekker, Lukas and van Der Harst, Pim and Blaauw, Yuri and Dichtl, Wolfgang and Senoner, Thomas and Hassink, Rutger J and Loh, Peter and others},
  journal={European Journal of Clinical Investigation},
  volume={55},
  pages={e14385},
  year={2025},
  publisher={Wiley Online Library}
}

@article{van2024automatic,
  title={Automatic triage of twelve-lead electrocardiograms using deep convolutional neural networks: a first implementation study},
  author={van de Leur, Rutger R and van Sleuwen, Meike TGM and Zwetsloot, Peter-Paul M and van der Harst, Pim and Doevendans, Pieter A and Hassink, Rutger J and van Es, Ren{\'e}},
  journal={European Heart Journal-Digital Health},
  volume={5},
  number={1},
  pages={89--96},
  year={2024},
  publisher={Oxford University Press US}
}

@article{mcgavigan2007localization,
  title={Localization of Accessory Pathways in the Wolff-Parkinson-White Pattern—Physician Versus Computer Interpretation of the Same Algorithm},
  author={McGAVIGAN, ANDREW D and Clark, Elaine and Quinn, F Russell and Rankin, Andrew C and Macfarlane, Peter W},
  journal={Pacing and clinical electrophysiology},
  volume={30},
  number={8},
  pages={998--1002},
  year={2007},
  publisher={Wiley Online Library}
}

@article{teixeira2016accuracy,
  title={Accuracy of the electrocardiogram in localizing the accessory pathway in patients with Wolff-Parkinson-White pattern},
  author={Teixeira, Carlos Manuel and Pereira, Telmo Ant{\'o}nio and Lebreiro, Ana Margarida and Carvalho, S{\'e}rgio Alexandre},
  journal={Arquivos brasileiros de cardiologia},
  volume={107},
  number={4},
  pages={331--338},
  year={2016},
  publisher={SciELO Brasil}
}

@article{kurath2022accuracy,
  title={Accuracy of algorithms predicting accessory pathway localization in pediatric patients with Wolff-Parkinson-White syndrome},
  author={Kurath-Koller, Stefan and Manninger, Martin and {\"O}ffl, Nathalie and K{\"o}stenberger, Martin and Sallmon, Hannes and Will, Joachim and Scherr, Daniel},
  journal={Children},
  volume={9},
  number={12},
  pages={1962},
  year={2022},
  publisher={MDPI}
}

@article{baek2020new,
  title={New algorithm for accessory pathway localization focused on screening septal pathways in pediatric patients with Wolff-Parkinson-White syndrome},
  author={Baek, Seung Min and Song, Mi Kyoung and Uhm, Jae-Sun and Kim, Gi Beom and Bae, Eun Jung},
  journal={Heart Rhythm},
  volume={17},
  number={12},
  pages={2172--2179},
  year={2020},
  publisher={Elsevier}
}

@article{li2019novel,
  title={A novel and simple algorithm using surface electrocardiogram that localizes accessory conduction pathway in Wolff-Parkinson-White syndrome in pediatric patients},
  author={Li, Hsing-Yuan and Chang, Shih-Lin and Chuang, Chi-Hsi and Lin, Ming-Chih and Lin, Yenn-Jiang and Lo, Li-Wei and Hu, Yu-Feng and Chung, Fa-Po and Chang, Yao-Ting and Chung, Chieh-Mao and others},
  journal={Acta Cardiologica Sinica},
  volume={35},
  number={5},
  pages={493},
  year={2019}
}

@article{wren2012accuracy,
  title={Accuracy of algorithms to predict accessory pathway location in children with Wolff--Parkinson--White syndrome},
  author={Wren, Christopher and Vogel, Melanie and Lord, Stephen and Abrams, Dominic and Bourke, John and Rees, Philip and Rosenthal, Eric},
  journal={Heart},
  volume={98},
  number={3},
  pages={202--206},
  year={2012},
  publisher={BMJ Publishing Group Ltd and British Cardiovascular Society}
}

@article{boersma2002accessory,
  title={Accessory pathway localization by QRS polarity in children with Wolff-Parkinson-White syndrome},
  author={Boersma, Lucas and Garc{\'\i}a-Moran, EMILIO and Mont, Llu{\'\i}s and Brugada, Josep},
  journal={Journal of cardiovascular electrophysiology},
  volume={13},
  number={12},
  pages={1222--1226},
  year={2002},
  publisher={Wiley Online Library}
}

@article{iturralde1996new,
  title={A new ECG algorithm for the localization of accessory pathways using only the polarity of the QRS complex},
  author={Iturralde, Pedro and Araya-Gomez, Vivien and Colin, Luis and Kershenovich, Sergio and de Micheli, Alfredo and Gonzalez-Hermosillo, J Antonio},
  journal={Journal of electrocardiology},
  volume={29},
  number={4},
  pages={289--299},
  year={1996},
  publisher={Elsevier}
}

@article{d1995fast,
  title={A fast and reliable algorithm to localize accessory pathways based on the polarity of the QRS complex on the surface ECG during sinus rhythm},
  author={d'Avila, Andre and Brugada, Josep and Skeberis, Vassilis and Andries, Erik and Sosa, Eduardo and Brugada, Pedro},
  journal={Pacing and Clinical Electrophysiology},
  volume={18},
  number={9},
  pages={1615--1627},
  year={1995},
  publisher={Wiley Online Library}
}

@article{chiang1995accurate,
  title={An accurate stepwise electrocardiographic algorithm for localization of accessory pathways in patients with Wolff-Parkinson-White syndrome from a comprehensive analysis of delta waves and R/S ratio during sinus rhythm},
  author={Chiang, Chern-En and Chen, Shih-Ann and Teo, Wee Siong and Tsai, Der-Shang and Wu, Tsu-Juey and Cheng, Chen-Chuan and Chiou, Chuen-Wang and Tai, Ching-Tai and Lee, Shih-Huang and Chen, Chung-Yin and others},
  journal={The American journal of cardiology},
  volume={76},
  number={1-2},
  pages={40--46},
  year={1995},
  publisher={Elsevier}
}

@article{xie1994localization,
  title={Localization of accessory pathways from the 12-lead electrocardiogram using a new algorithm},
  author={Xie, Baiyan and Heald, Spencer C and Bashir, Yaver and Katritsis, Demosthenes and Murgatroyd, Francis D and Camm, A John and Rowland, Edward and Ward, David E},
  journal={The American journal of cardiology},
  volume={74},
  number={2},
  pages={161--165},
  year={1994},
  publisher={Elsevier}
}

@article{milstein1987algorithm,
  title={An algorithm for the electrocardiographic localization of accessory pathways in the Wolff-Parkinson-White syndrome},
  author={Milstein, Simon and Sharma, Arjun D and Guiraudon, Gerard M and Klein, George J},
  journal={Pacing and Clinical Electrophysiology},
  volume={10},
  number={3},
  pages={555--563},
  year={1987},
  publisher={Wiley Online Library}
}

@article{khalaph2025novel,
  title={A novel ECG algorithm for accurate localization of manifest accessory pathways in both children and adults: SMART-WPW},
  author={Khalaph, Moneeb and Trajkovska, Nadica and Didenko, Maxim and Braun, Martin and Imnadze, Guram and Akkaya, Ersan and Fink, Thomas and Lucas, Philipp and Sciacca, Vanessa and Beyer, Sebastian and others},
  journal={Heart Rhythm},
  year={2025},
  publisher={Elsevier}
}

@article{fujino2020clinical,
  title={Clinical characteristics of challenging catheter ablation procedures in patients with WPW syndrome: A 10 year single-center experience},
  author={Fujino, Tadashi and De Ruvo, Ermengildo and Grieco, Domenico and Scar{\'a}, Antonio and Borrelli, Alessio and De Luca, Lucia and Panuccio, Marco and Fagagnini, Alessandro and Bruni, Giuseppe and Sciarra, Luigi and others},
  journal={Journal of Cardiology},
  volume={76},
  number={4},
  pages={420--426},
  year={2020},
  publisher={Elsevier}
}

@article{belyadi2021unsupervised,
  title={Unsupervised machine learning: clustering algorithms},
  author={Belyadi, H and Haghighat, A},
  journal={Machine Learning Guide for Oil and Gas Using Python},
  pages={125--168},
  year={2021},
  publisher={Elsevier Amsterdam, The Netherlands}
}

@article{sherdia2023success,
  title={The success rate of radiofrequency catheter ablation in Wolff-Parkinson-White-Syndrome patients: A systematic review and meta-analysis},
  author={Sherdia, Abdelrahman Farag Ibrahim Ali and Abdelaal, Shadi Alaa and Hasan, Mohammed Tarek and Elsayed, Esraa and Mare'y, Mohamed and Nawar, Asmaa Ahmed and Abdelsalam, Alaa and Abdelgader, Mujtaba Zakria and Adam, Alameen and Abozaid, Mohamed},
  journal={Indian Heart Journal},
  volume={75},
  number={2},
  pages={98--107},
  year={2023},
  publisher={Elsevier}
}

@article{schiavone2024pre,
  title={Pre-Excited Atrial Fibrillation in Wolff-Parkinson-White (WPW) Syndrome: A Case Report and a Review of the Literature},
  author={Schiavone, Marco and Filtz, Annalisa and Gasperetti, Alessio and Zhang, Xiaodong and Forleo, Giovanni B and Santangeli, Pasquale and Di Biase, Luigi},
  journal={Reviews in Cardiovascular Medicine},
  volume={25},
  number={4},
  pages={125},
  year={2024}
}

@article{pappone2012risk,
  title={Risk of malignant arrhythmias in initially symptomatic patients with Wolff-Parkinson-White syndrome: results of a prospective long-term electrophysiological follow-up study},
  author={Pappone, Carlo and Vicedomini, Gabriele and Manguso, Francesco and Baldi, Mario and Pappone, Alessia and Petretta, Andrea and Vitale, Raffaele and Saviano, Massimo and Ciaccio, Cristiano and Giannelli, Luigi and others},
  journal={Circulation},
  volume={125},
  number={5},
  pages={661--668},
  year={2012},
  publisher={Lippincott Williams \& Wilkins Hagerstown, MD}
}

@article{paerregaard2023wolff,
  title={The Wolff--Parkinson--White pattern in neonates: results from a large population-based cohort study},
  author={P{\ae}rregaard, Maria Munk and Hartmann, Joachim and Sillesen, Anne-Sophie and Pihl, Christian and Dannesbo, Sofie and Kock, Thilde Olivia and Pietersen, Adrian and Raja, Anna Axelsson and Iversen, Kasper Karmark and Bundgaard, Henning and others},
  journal={Europace},
  volume={25},
  number={7},
  pages={euad165},
  year={2023},
  publisher={Oxford University Press US}
}

@article{elendu2025risk,
  title={Risk stratification and management of arrhythmias in patients with Wolff--Parkinson--White syndrome},
  author={Elendu, Chukwuka and Babarinde, Festus O and Babatunde, Olusola D and Babawale, Emmanuel A and Hassan, Jemilah I and Ikeji, Victor I and Oshin, Boluwatife D and Nwabueze, Adaugo and Ngozi-Ibeh, Jide K and Chukwu, Christopher and others},
  journal={Annals of Medicine and Surgery},
  volume={87},
  number={5},
  pages={2702--2717},
  year={2025},
  publisher={LWW}
}

@article{sapra2020wolff,
  title={Wolff-Parkinson-White Syndrome: a master of disguise},
  author={Sapra, Amit and Albers, Janet and Bhandari, Priyanka and Davis, Dean and Ranjit, Eukesh},
  journal={Cureus},
  volume={12},
  number={6},
  year={2020},
  publisher={Cureus}
}

@article{vuatuașescu2024wolf,
  title={Wolf--Parkinson--White Syndrome: Diagnosis, Risk Assessment, and Therapy—An Update},
  author={V{\u{a}}t{\u{a}}\d{s}escu, Radu Gabriel and Paja, Cosmina Steliana and \d{S}u\d{s}, Ioana and Cainap, Simona and Moisa, \d{S}tefana Mar{\'\i}a and Cintez{\u{a}}, Eliza Elena},
  journal={Diagnostics},
  volume={14},
  number={3},
  pages={296},
  year={2024},
  publisher={MDPI}
}

@article{arrudaDevelopmentValidationECG1998,
  title = {Development and Validation of an {{ECG}} Algorithm for Identifying Accessory Pathway Ablation Site in {{Wolff-Parkinson-White}} Syndrome},
  author = {Arruda, M. S. and McClelland, J. H. and Wang, X. and Beckman, K. J. and Widman, L. E. and Gonzalez, M. D. and Nakagawa, H. and Lazzara, R. and Jackman, W. M.},
  year = {1998},
  month = jan,
  journal = {Journal of Cardiovascular Electrophysiology},
  volume = {9},
  number = {1},
  pages = {2--12},
  issn = {1045-3873},
  doi = {10.1111/j.1540-8167.1998.tb00861.x},
  langid = {english},
  pmid = {9475572}
}

@article{ayano2024interpretable,
  title={Interpretable Hybrid Multichannel Deep Learning Model for Heart Disease Classification Using 12-leads ECG Signal},
  author={Ayano, Yehualashet Megersa and Schwenker, Friedhelm and Dufera, Bisrat Derebssa and Debelee, Taye Girma and Ejegu, Yitagesu Getachew},
  journal={IEEE Access},
  year={2024},
  publisher={IEEE}
}

@article{ayanoInterpretableMachineLearning2022,
  title = {Interpretable {{Machine Learning Techniques}} in {{ECG-Based Heart Disease Classification}}: {{A Systematic Review}}},
  shorttitle = {Interpretable {{Machine Learning Techniques}} in {{ECG-Based Heart Disease Classification}}},
  author = {Ayano, Yehualashet Megersa and Schwenker, Friedhelm and Dufera, Bisrat Derebssa and Debelee, Taye Girma},
  year = {2022},
  month = dec,
  journal = {Diagnostics (Basel, Switzerland)},
  volume = {13},
  number = {1},
  pages = {111},
  issn = {2075-4418},
  doi = {10.3390/diagnostics13010111},
  langid = {english},
  pmcid = {PMC9818170},
  pmid = {36611403}
}

@book{bishopPatternRecognitionMachine2006,
  title = {Pattern Recognition and Machine Learning},
  author = {Bishop, Christopher M.},
  year = {2006},
  series = {Information Science and Statistics},
  publisher = {Springer},
  address = {New York},
  isbn = {978-0-387-31073-2},
  langid = {english},
  lccn = {006.4}
}

@article{elhamritiEASYWPWNovelECGalgorithm2023,
  title = {{{EASY-WPW}}: A Novel {{ECG-algorithm}} for Easy and Reliable Localization of Manifest Accessory Pathways in Children and Adults},
  shorttitle = {{{EASY-WPW}}},
  author = {El Hamriti, Mustapha and Braun, Martin and Molatta, Stephan and Imnadze, Guram and Khalaph, Moneeb and Lucas, Philipp and Nolting, Julia Kathinka and Isgandarova, Khuraman and Sciacca, Vanessa and Fink, Thomas and Bergau, Leonard and Sohns, Christian and Kiuchi, Kunihiko and Nishimori, Makoto and Heeger, Christian-Hendrik and Borlich, Martin and Shin, Dong-In and Busch, Sonia and Guckel, Denise and Sommer, Philipp},
  year = {2023},
  month = feb,
  journal = {Europace: European Pacing, Arrhythmias, and Cardiac Electrophysiology: Journal of the Working Groups on Cardiac Pacing, Arrhythmias, and Cardiac Cellular Electrophysiology of the European Society of Cardiology},
  volume = {25},
  number = {2},
  pages = {600--609},
  issn = {1532-2092},
  doi = {10.1093/europace/euac216},
  langid = {english},
  pmcid = {PMC9935024},
  pmid = {36504238}
}

@article{fananapazirImportancePreexcitedQRS1990,
  title = {Importance of Preexcited {{QRS}} Morphology during Induced Atrial Fibrillation to the Diagnosis and Localization of Multiple Accessory Pathways.},
  author = {Fananapazir, L and German, L D and Gallagher, J J and Lowe, J E and Prystowsky, E N},
  year = {1990},
  month = feb,
  journal = {Circulation},
  volume = {81},
  number = {2},
  pages = {578--585},
  publisher = {American Heart Association},
  doi = {10.1161/01.CIR.81.2.578},
  urldate = {2025-05-07}
}

@article{fawazDeepLearningTime2019,
  title = {Deep Learning for Time Series Classification: A Review},
  shorttitle = {Deep Learning for Time Series Classification},
  author = {Fawaz, Hassan Ismail and Forestier, Germain and Weber, Jonathan and Idoumghar, Lhassane and Muller, Pierre-Alain},
  year = {2019},
  month = jul,
  journal = {Data Mining and Knowledge Discovery},
  volume = {33},
  number = {4},
  eprint = {1809.04356},
  primaryclass = {cs},
  pages = {917--963},
  issn = {1384-5810, 1573-756X},
  doi = {10.1007/s10618-019-00619-1},
  urldate = {2025-04-30},
  archiveprefix = {arXiv},
  langid = {english}
}

@article{gilletteComputationalStudyInfluence2025,
  title = {A Computational Study on the Influence of Antegrade Accessory Pathway Location on the 12-Lead Electrocardiogram in {{Wolff}}--{{Parkinson}}--{{White}} Syndrome},
  author = {Gillette, Karli and Winkler, Benjamin and {Kurath-Koller}, Stefan and Scherr, Daniel and Vigmond, Edward J and B{\"a}r, Markus and Plank, Gernot},
  year = {2025},
  month = feb,
  journal = {Europace},
  volume = {27},
  number = {2},
  pages = {euae223},
  issn = {1099-5129, 1532-2092},
  doi = {10.1093/europace/euae223},
  urldate = {2025-04-30},
  copyright = {https://creativecommons.org/licenses/by/4.0/},
  langid = {english}
}

@inproceedings{glorotDeepSparseRectifier2011,
  title = {Deep {{Sparse Rectifier Neural Networks}}},
  booktitle = {Proceedings of the {{Fourteenth International Conference}} on {{Artificial Intelligence}} and {{Statistics}}},
  author = {Glorot, Xavier and Bordes, Antoine and Bengio, Yoshua},
  year = {2011},
  month = jun,
  pages = {315--323},
  publisher = {{JMLR Workshop and Conference Proceedings}},
  issn = {1938-7228},
  urldate = {2025-05-07},
  langid = {english}
}

@article{haissaguerreElectrocardiographicCharacteristicsCatheter1994,
  title = {Electrocardiographic Characteristics and Catheter Ablation of Parahissian Accessory Pathways},
  author = {Haissaguerre, M. and Marcus, F. and Poquet, F. and Gencel, L. and Le M{\'e}tayer, P. and Cl{\'e}menty, J.},
  year = {1994},
  month = sep,
  journal = {Circulation},
  volume = {90},
  number = {3},
  pages = {1124--1128},
  issn = {0009-7322},
  doi = {10.1161/01.cir.90.3.1124},
  langid = {english},
  pmid = {8087922}
}

@book{haykinNeuralNetworksLearning2009a,
  title = {Neural {{Networks}} and {{Learning Machines}}},
  author = {Haykin, Simon S.},
  year = {2009},
  publisher = {Prentice Hall},
  googlebooks = {K7P36lKzI\_QC},
  isbn = {978-0-13-147139-9},
  langid = {english}
}

@misc{ioffeBatchNormalizationAccelerating2015,
  title = {Batch {{Normalization}}: {{Accelerating Deep Network Training}} by {{Reducing Internal Covariate Shift}}},
  shorttitle = {Batch {{Normalization}}},
  author = {Ioffe, Sergey and Szegedy, Christian},
  year = {2015},
  month = mar,
  number = {arXiv:1502.03167},
  eprint = {1502.03167},
  primaryclass = {cs},
  publisher = {arXiv},
  doi = {10.48550/arXiv.1502.03167},
  urldate = {2025-05-07},
  archiveprefix = {arXiv}
}

@article{jahmunahExplainableDetectionMyocardial2022,
  title = {Explainable Detection of Myocardial Infarction Using Deep Learning Models with {{Grad-CAM}} Technique on {{ECG}} Signals},
  author = {Jahmunah, V. and Ng, E.Y.K. and Tan, Ru-San and Oh, Shu Lih and Acharya, U Rajendra},
  year = {2022},
  month = jul,
  journal = {Computers in Biology and Medicine},
  volume = {146},
  pages = {105550},
  issn = {00104825},
  doi = {10.1016/j.compbiomed.2022.105550},
  urldate = {2025-04-30},
  langid = {english}
}

@inproceedings{jonesImprovingECGClassification2020,
  title = {Improving {{ECG Classification Interpretability}} Using {{Saliency Maps}}},
  booktitle = {2020 {{IEEE}} 20th {{International Conference}} on {{Bioinformatics}} and {{Bioengineering}} ({{BIBE}})},
  author = {Jones, Yola and Deligianni, Fani and Dalton, Jeff},
  year = {2020},
  month = oct,
  eprint = {2201.04070},
  primaryclass = {eess},
  pages = {675--682},
  doi = {10.1109/BIBE50027.2020.00114},
  urldate = {2025-05-08},
  archiveprefix = {arXiv}
}

@article{middlehurst2024bake,
  title={Bake off redux: a review and experimental evaluation of recent time series classification algorithms},
  author={Middlehurst, Matthew and Sch{\"a}fer, Patrick and Bagnall, Anthony},
  journal={Data Mining and Knowledge Discovery},
  volume={38},
  number={4},
  pages={1958--2031},
  year={2024},
  publisher={Springer}
}

@misc{leLightX3ECGLightweightEXplainable2022,
  title = {{{LightX3ECG}}: {{A Lightweight}} and {{eXplainable Deep Learning System}} for 3-Lead {{Electrocardiogram Classification}}},
  shorttitle = {{{LightX3ECG}}},
  author = {Le, Khiem H. and Pham, Hieu H. and Nguyen, Thao BT and Nguyen, Tu A. and Thanh, Tien N. and Do, Cuong D.},
  year = {2022},
  month = jul,
  number = {arXiv:2207.12381},
  eprint = {2207.12381},
  primaryclass = {cs},
  publisher = {arXiv},
  doi = {10.48550/arXiv.2207.12381},
  urldate = {2025-05-08},
  archiveprefix = {arXiv}
}

@misc{linNetworkNetwork2014,
  title = {Network {{In Network}}},
  author = {Lin, Min and Chen, Qiang and Yan, Shuicheng},
  year = {2014},
  month = mar,
  number = {arXiv:1312.4400},
  eprint = {1312.4400},
  primaryclass = {cs},
  publisher = {arXiv},
  doi = {10.48550/arXiv.1312.4400},
  urldate = {2025-05-07},
  archiveprefix = {arXiv}
}

@article{muller2007dynamic,
  title={Dynamic time warping},
  author={M{\"u}ller, Meinard},
  journal={Information retrieval for music and motion},
  pages={69--84},
  year={2007},
  publisher={Springer}
}

@article{izakian2015fuzzy,
  title={Fuzzy clustering of time series data using dynamic time warping distance},
  author={Izakian, Hesam and Pedrycz, Witold and Jamal, Iqbal},
  journal={Engineering Applications of Artificial Intelligence},
  volume={39},
  pages={235--244},
  year={2015},
  publisher={Elsevier}
}

@article{nishimoriAccessoryPathwayAnalysis2021,
  title = {Accessory Pathway Analysis Using a Multimodal Deep Learning Model},
  author = {Nishimori, Makoto and Kiuchi, Kunihiko and Nishimura, Kunihiro and Kusano, Kengo and Yoshida, Akihiro and Adachi, Kazumasa and Hirayama, Yasutaka and Miyazaki, Yuichiro and Fujiwara, Ryudo and Sommer, Philipp and El Hamriti, Mustapha and Imada, Hiroshi and Takemoto, Makoto and Takami, Mitsuru and Shinohara, Masakazu and Toh, Ryuji and Fukuzawa, Koji and Hirata, Ken-Ichi},
  year = {2021},
  month = apr,
  journal = {Scientific Reports},
  volume = {11},
  number = {1},
  pages = {8045},
  issn = {2045-2322},
  doi = {10.1038/s41598-021-87631-y},
  langid = {english},
  pmcid = {PMC8044112},
  pmid = {33850245}
}

@article{pambrunMaximalPreExcitationBased2018,
  title = {Maximal {{Pre-Excitation Based Algorithm}} for {{Localization}} of {{Manifest Accessory Pathways}} in {{Adults}}},
  author = {Pambrun, Thomas and El Bouazzaoui, Rim and Combes, Nicolas and Combes, St{\'e}phane and Sousa, Pedro and Le Bloa, Mathieu and Massoulli{\'e}, Gr{\'e}goire and Cheniti, Ghassen and Martin, Ruairidh and Pillois, Xavier and Duchateau, Josselin and Sacher, Fr{\'e}d{\'e}ric and Hocini, M{\'e}l{\`e}ze and Ja{\"i}s, Pierre and Derval, Nicolas and Bortone, Agust{\'i}n and Boveda, Serge and Denis, Arnaud and Ha{\"i}ssaguerre, Michel and Albenque, Jean-Paul},
  year = {2018},
  month = aug,
  journal = {JACC Clinical electrophysiology},
  volume = {4},
  number = {8},
  pages = {1052--1061},
  issn = {2405-5018},
  doi = {10.1016/j.jacep.2018.03.018},
  urldate = {2025-05-07},
  langid = {english},
  pmid = {30139487}
}

@article{ramirezArtSelectingECG2024,
  title = {The Art of Selecting the {{ECG}} Input in Neural Networks to Classify Heart Diseases: A Dual Focus on Maximizing Information and Reducing Redundancy},
  shorttitle = {The Art of Selecting the {{ECG}} Input in Neural Networks to Classify Heart Diseases},
  author = {Ramirez, Elisa and {Ruiperez-Campillo}, Samuel and {Casado-Arroyo}, Ruben and Merino, Jos{\'e} Luis and Vogt, Julia E. and Castells, Francisco and Millet, Jos{\'e}},
  year = {2024},
  month = oct,
  journal = {Frontiers in Physiology},
  volume = {15},
  publisher = {Frontiers},
  issn = {1664-042X},
  doi = {10.3389/fphys.2024.1452829},
  urldate = {2025-05-08},
  langid = {english}
}

@article{rosafalcoOnlineStructuralHealth2021,
  title = {Online Structural Health Monitoring by Model Order Reduction and Deep Learning Algorithms},
  author = {Rosafalco, Luca and Torzoni, Matteo and Manzoni, Andrea and Mariani, Stefano and Corigliano, Alberto},
  year = {2021},
  month = oct,
  journal = {Computers \& Structures},
  volume = {255},
  pages = {106604},
  issn = {00457949},
  doi = {10.1016/j.compstruc.2021.106604},
  urldate = {2025-04-30},
  langid = {english}
}

@incollection{samekExplainableArtificialIntelligence2019,
  title = {Towards {{Explainable Artificial Intelligence}}},
  author = {Samek, Wojciech and M{\"u}ller, Klaus-Robert},
  year = {2019},
  volume = {11700},
  eprint = {1909.12072},
  primaryclass = {cs},
  pages = {5--22},
  doi = {10.1007/978-3-030-28954-6_1},
  urldate = {2025-05-08},
  archiveprefix = {arXiv}
}

@article{selvarajuGradCAMVisualExplanations2020,
  title = {Grad-{{CAM}}: {{Visual Explanations}} from {{Deep Networks}} via {{Gradient-based Localization}}},
  shorttitle = {Grad-{{CAM}}},
  author = {Selvaraju, Ramprasaath R. and Cogswell, Michael and Das, Abhishek and Vedantam, Ramakrishna and Parikh, Devi and Batra, Dhruv},
  year = {2020},
  month = feb,
  journal = {International Journal of Computer Vision},
  volume = {128},
  number = {2},
  eprint = {1610.02391},
  primaryclass = {cs},
  pages = {336--359},
  issn = {0920-5691, 1573-1405},
  doi = {10.1007/s11263-019-01228-7},
  urldate = {2025-04-30},
  archiveprefix = {arXiv},
  langid = {english}
}

@article{senonerIdentifyingLocationAccessory2021,
  title = {Identifying the {{Location}} of an {{Accessory Pathway}} in {{Pre-Excitation Syndromes Using}} an {{Artificial Intelligence-Based Algorithm}}},
  author = {Senoner, Thomas and Pfeifer, Bernhard and Barbieri, Fabian and Adukauskaite, Agne and Dichtl, Wolfgang and Bauer, Axel and Hintringer, Florian},
  year = {2021},
  month = sep,
  journal = {Journal of Clinical Medicine},
  volume = {10},
  number = {19},
  pages = {4394},
  issn = {2077-0383},
  doi = {10.3390/jcm10194394},
  langid = {english},
  pmcid = {PMC8509837},
  pmid = {34640411}
}

@article{siontisSaliencyMapsProvide2023,
  title = {Saliency Maps Provide Insights into Artificial Intelligence-Based Electrocardiography Models for Detecting Hypertrophic Cardiomyopathy},
  author = {Siontis, Konstantinos C. and Su{\'a}rez, Abraham B{\'a}ez and Sehrawat, Ojasav and Ackerman, Michael J. and Attia, Zachi I. and Friedman, Paul A. and Noseworthy, Peter A. and Maanja, Maren},
  year = {2023},
  journal = {Journal of Electrocardiology},
  volume = {81},
  pages = {286--291},
  issn = {1532-8430},
  doi = {10.1016/j.jelectrocard.2023.07.002},
  langid = {english},
  pmid = {37599145}
}

@misc{springenbergStrivingSimplicityAll2015,
  title = {Striving for {{Simplicity}}: {{The All Convolutional Net}}},
  shorttitle = {Striving for {{Simplicity}}},
  author = {Springenberg, Jost Tobias and Dosovitskiy, Alexey and Brox, Thomas and Riedmiller, Martin},
  year = {2015},
  month = apr,
  number = {arXiv:1412.6806},
  eprint = {1412.6806},
  primaryclass = {cs},
  publisher = {arXiv},
  doi = {10.48550/arXiv.1412.6806},
  urldate = {2025-04-30},
  archiveprefix = {arXiv},
  langid = {english}
}

@article{suhVisualInterpretationDeep2024,
  title = {Visual Interpretation of Deep Learning Model in {{ECG}} Classification: {{A}} Comprehensive Evaluation of Feature Attribution Methods},
  shorttitle = {Visual Interpretation of Deep Learning Model in {{ECG}} Classification},
  author = {Suh, Jangwon and Kim, Jimyeong and Kwon, Soonil and Jung, Euna and Ahn, Hyo-Jeong and Lee, Kyung-Yeon and Choi, Eue-Keun and Rhee, Wonjong},
  year = {2024},
  month = nov,
  journal = {Computers in Biology and Medicine},
  volume = {182},
  pages = {109088},
  issn = {00104825},
  doi = {10.1016/j.compbiomed.2024.109088},
  urldate = {2025-04-30},
  langid = {english}
}

@article{theisslerExplainableAITime2022,
  title = {Explainable {{AI}} for {{Time Series Classification}}: {{A Review}}, {{Taxonomy}} and {{Research Directions}}},
  shorttitle = {Explainable {{AI}} for {{Time Series Classification}}},
  author = {Theissler, Andreas and Spinnato, Francesco and Schlegel, Udo and Guidotti, Riccardo},
  year = {2022},
  journal = {IEEE Access},
  volume = {10},
  pages = {100700--100724},
  issn = {2169-3536},
  doi = {10.1109/ACCESS.2022.3207765},
  urldate = {2025-04-30},
  copyright = {https://creativecommons.org/licenses/by/4.0/legalcode},
  langid = {english}
}

@article{xie2024intelligent,
  title={Intelligent Analysis and Heartbeat Saliency Map Representation of Postoperative Atrial Fibrillation Recurrence Based on Mobile Single-lead Electrocardiogram},
  author={Xie, Yushan and Zhu, Huaiyu and Chen, Laite and Chen, Wensheng and Jiang, Chenyang and Pan, Yun},
  journal={IEEE Transactions on Instrumentation and Measurement},
  year={2024},
  publisher={IEEE}
}

@inproceedings{jones2020improving,
  title={Improving ECG classification interpretability using saliency maps},
  author={Jones, Yola and Deligianni, Fani and Dalton, Jeff},
  booktitle={2020 IEEE 20th International Conference on Bioinformatics and Bioengineering (BIBE)},
  pages={675--682},
  year={2020},
  organization={IEEE}
}

@article{agrawal2022ecg,
  title={ECG-iCOVIDNet: Interpretable AI model to identify changes in the ECG signals of post-COVID subjects},
  author={Agrawal, Amulya and Chauhan, Aniket and Shetty, Manu Kumar and Gupta, Mohit D and Gupta, Anubha and others},
  journal={Computers in Biology and Medicine},
  volume={146},
  pages={105540},
  year={2022},
  publisher={Elsevier}
}

@article{vandeleurDiscoveringVisualizingDiseaseSpecific2021,
  title = {Discovering and {{Visualizing Disease-Specific Electrocardiogram Features Using Deep Learning}}},
  author = {{van de Leur}, Rutger R. and Taha, Karim and Bos, Max N. and {van der Heijden}, Jeroen F. and Gupta, Deepak and Cramer, Maarten J. and Hassink, Rutger J. and {van der Harst}, Pim and Doevendans, Pieter A. and Asselbergs, Folkert W. and {van Es}, Ren{\'e}},
  year = {2021},
  month = jan,
  journal = {Circulation. Arrhythmia and Electrophysiology},
  volume = {14},
  number = {2},
  pages = {e009056},
  issn = {1941-3149},
  doi = {10.1161/CIRCEP.120.009056},
  urldate = {2025-05-08},
  pmcid = {PMC7892204},
  pmid = {33401921}
}

@article{vandeleurECGonlyExplainableDeep2024,
  title = {{{ECG-only}} Explainable Deep Learning Algorithm Predicts the Risk for Malignant Ventricular Arrhythmia in Phospholamban Cardiomyopathy},
  author = {{van de Leur}, Rutger R. and {de Brouwer}, Remco and Bleijendaal, Hidde and Verstraelen, Tom E. and Mahmoud, Belend and {Perez-Matos}, Ana and Dickhoff, Cathelijne and Schoonderwoerd, Bas A. and Germans, Tjeerd and Houweling, Arjan and {van der Zwaag}, Paul A. and Cox, Moniek G. P. J. and {Peter van Tintelen}, J. and Te Riele, Anneline S. J. M. and {van den Berg}, Maarten P. and Wilde, Arthur A. M. and Doevendans, Pieter A. and {de Boer}, Rudolf A. and {van Es}, Ren{\'e}},
  year = {2024},
  month = jul,
  journal = {Heart Rhythm},
  volume = {21},
  number = {7},
  pages = {1102--1112},
  issn = {1556-3871},
  doi = {10.1016/j.hrthm.2024.02.038},
  langid = {english},
  pmid = {38403235}
}

@article{yahyazadehNovelFeatureExtraction2024,
  title = {A Novel Feature Extraction Method for the Localization of Accessory Pathways in Patients with {{Wolff-Parkinson-White}} Syndrome},
  author = {Yahyazadeh, Sakineh and Jafarnia Dabanloo, Nader and Nasrabadi, Ali Motie and Ghorbani Sharif, Alireza},
  year = {2024},
  month = feb,
  journal = {Biomedical Signal Processing and Control},
  volume = {88},
  pages = {105640},
  issn = {17468094},
  doi = {10.1016/j.bspc.2023.105640},
  urldate = {2025-04-30},
  langid = {english}
}
\appendix
\section{Selected values for hyperparameters}\label{sec:hyperparameters} \label{sec:hyperparameters}
The three \gls{fcn} architectures were trained with the hyperparameter settings summarized in Table \ref{tab:hyperparameters}, resulting in 1161016, 531000, 3996120 trainable parameters, respectively. The best performance on the test set was yielded by the \gls{fcn} with the lowest complexity and, therefore, the least expensive to train.
\begin{table}[ht]
\centering
\renewcommand{\arraystretch}{1.2}
\begin{tabular}{c|c|ccc}
\textbf{Layer} & \textbf{Hyperparameter} & \textbf{FCN 2.2.1} & \textbf{FCN 2.2.2} & \textbf{FCN 2.2.3} \\ \hline
\multirow{3}{*}{Layer 1} 
  & Num. filters & \cellcolor{LightGreen}96  & \cellcolor{LightGreen}96  & \cellcolor{LightGreen}128 \\
  & Kernel size  & \cellcolor{LightGreen}100   & \cellcolor{LightGreen}9 & \cellcolor{LightGreen}9   \\
  & Stride       & \cellcolor{LightGreen}10  & \cellcolor{LightYellow}1  & \cellcolor{LightYellow}1   \\ \hline
\multirow{3}{*}{Layer 2} 
  & Num. filters & \cellcolor{LightGreen}256 & \cellcolor{LightGreen}256 & \cellcolor{LightGreen}192 \\
  & Kernel size  & \cellcolor{LightGreen}20   & \cellcolor{LightGreen}9  & \cellcolor{LightGreen}9   \\
  & Stride       & \cellcolor{LightGreen}1  & \cellcolor{LightYellow}1   & \cellcolor{LightYellow}1   \\ \hline
\multirow{3}{*}{Layer 3} 
  & Num. filters & \cellcolor{LightGreen}128 & \cellcolor{LightGreen}128 & \cellcolor{LightGreen}128 \\
  & Kernel size  & \cellcolor{LightGreen}20   & \cellcolor{LightGreen}9  & \cellcolor{LightGreen}9   \\
  & Stride       & \cellcolor{LightGreen}2  & \cellcolor{LightYellow}1   & \cellcolor{LightYellow}1   \\
\end{tabular}
\caption{Hyperparameters of the three FCN variants for each layer. Entries in green refer to tuned values, while yellow highlights the prescribed stride hyperparameter.}
\label{tab:hyperparameters}
\end{table}

\section{Relationship between dataset regions and \gls{dt} regions}\label{sec:appendix_regions}


Given the differing definitions of anatomical regions within the ventricles used in \gls{ecg} classification, we compared our \gls{dl}-based model with classical \gls{dt} approaches by regrouping our regions to align with those defined by the \gls{dt} algorithms. Specifically, our 24 reference regions were mapped to the EASY-WPW algorithm \cite{elhamritiEASYWPWNovelECGalgorithm2023}, as shown in Table \ref{tab:easy-wpw}, and to the \gls{dt} algorithm proposed by Arruda \emph{et al.} \cite{arrudaDevelopmentValidationECG1998}, as shown in Table \ref{tab:arruda}. The mapping was manually constructed based on the descriptions provided in the reference works for EASY-WPW and Arruda -- positioning the ventricles in a $60\deg$ left anterior oblique view and following the authors reported instructions. Since our 24 regions include a subdivision along the longitudinal ventricular axis (apico-basal), rather than only a rotational subdivision as in the \gls{dt} algorithms, the mapping was defined based on the superior regions, i.e., those closest to the ventricular base. The inferior regions were then grouped with their corresponding superior regions according to rotational position.

\begin{figure}[htbp]
    \centering
    \includegraphics[width=\linewidth]{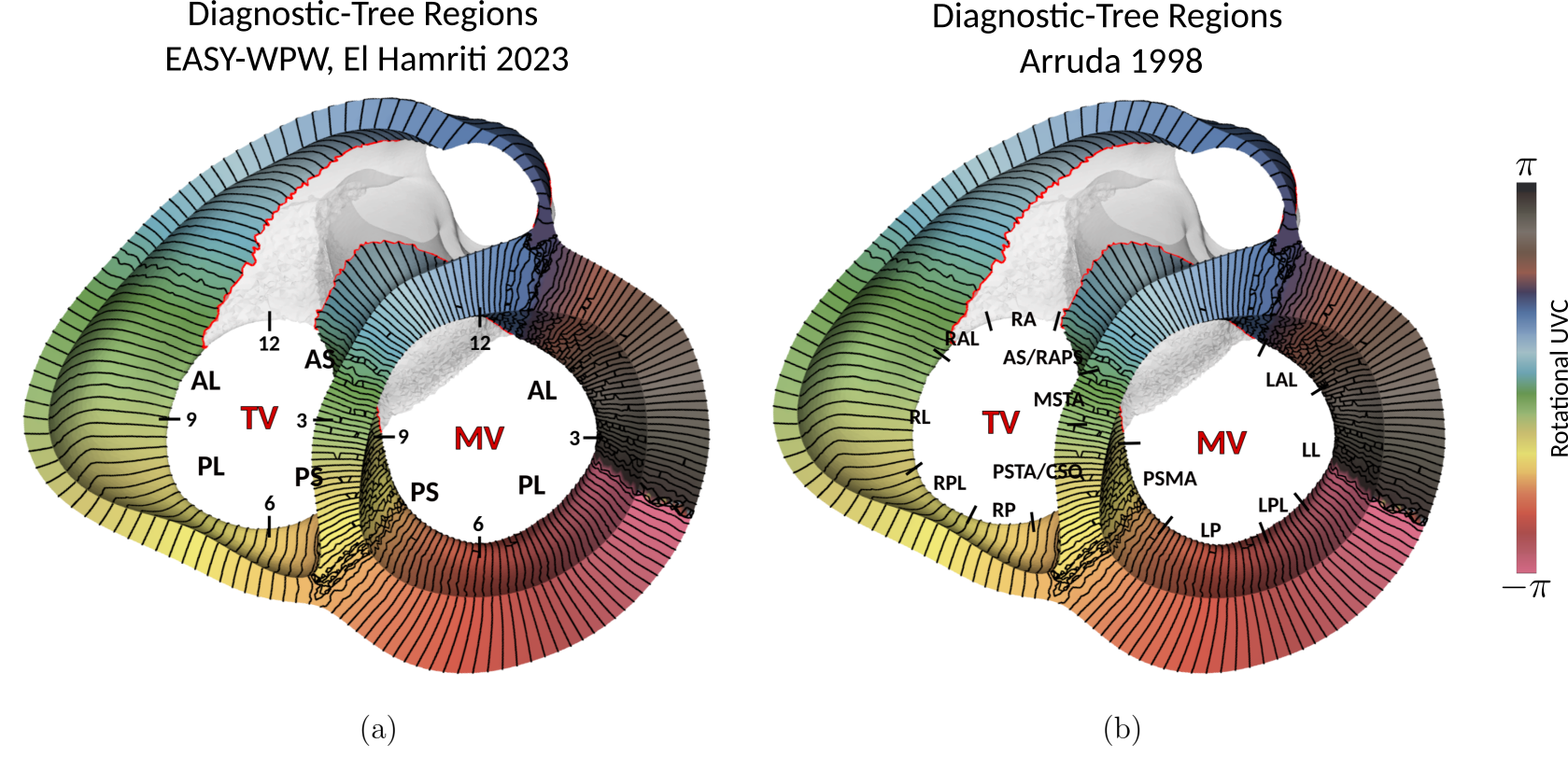}
    \caption{\gls{dt} region subdivision according to EASY-WPW \cite{elhamritiEASYWPWNovelECGalgorithm2023} algorithm (a) and Arruda  \cite{arrudaDevelopmentValidationECG1998} algorithm (b) on our ventricular model represented in a $60\deg$ left anterior oblique view.  The colors represent the distribution of the rotational \gls{uvc}.}
    \label{fig:diagnostic_regions}
\end{figure}

A random subset of 75 \glspl{ecg} from the test set was selected for classification using the EASY-WPW and Arruda \gls{dt} algorithms. The corresponding ground-truth \gls{ecg} labels, based on the \gls{dt} region subdivisions, were automatically derived using the \gls{uvc} framework. For each \gls{dt} region, a corresponding \gls{uvc} boundary interval was defined, and any \gls{ecg} generated by an \gls{ap} located within that interval was assigned to the corresponding \gls{dt} region. A graphical representation of this \gls{uvc}-based mapping is provided in Figure~\ref{fig:diagnostic_regions}.


\begin{table}[htbp]
\centering
\renewcommand{\arraystretch}{1.2}
\begin{tabular}{c|c}
\textbf{\gls{dt} region} & \textbf{Dataset regions} \\ \hline
MV-AL & $\{3, 9, 4, 10, 5, 11\}$ \\ \hline
MV-PL & $\{0, 6, 1, 7, 5, 11\}$ \\ \hline
MV-PS &  $\{1, 7, 2, 8\}$\\ \hline
TV-AL & $\{13, 19, 14, 20, 15, 21, 16, 22, 17, 23\}$\\ \hline
TV-PL & $\{12, 18, 13, 19\}$\\ \hline
TV-PS & $\{12, 18, 1, 7, 2, 8\}$\\ \hline
TV-AS & $\{17, 23, 2, 8, 3, 9, 4, 10\}$\\
\end{tabular}
\caption{Map between our 24 anatomical regions and the 7 regions used by the EASY-WPW algorithm presented in \cite{elhamritiEASYWPWNovelECGalgorithm2023}.}
\label{tab:easy-wpw}
\end{table}

\begin{table}[htbp]
\centering
\renewcommand{\arraystretch}{1.2}
\begin{tabular}{c|c}
\textbf{\gls{dt} region} & \textbf{Dataset regions} \\ \hline
LAL & $\{4, 10, 5, 11\}$ \\ \hline
LL & $\{0, 6, 5, 11\}$ \\ \hline
LP &  $\{1, 7, 0, 6\}$\\ \hline
LPL & $\{0, 6\}$\\ \hline
PSMA & $\{1, 7, 2, 8\}$\\ \hline
PSTA & $\{12, 18, 1, 7, 2, 8\}$\\ \hline
MSTA & $\{2, 8\}$\\\hline
AS & $\{2, 8, 3, 9\}$\\\hline
RA & $\{3, 9, 4, 10, 16, 22, 17, 23\}$\\\hline
RP & $\{12, 18\}$\\\hline
RPL & $\{12, 18, 13, 19\}$\\\hline
RL & $\{13, 19, 14, 20\}$\\\hline
RAL & $\{14, 20, 15, 21, 16, 22\}$
\end{tabular}
\caption{Map between our 24 anatomical regions and the 10 regions used by the \gls{dt} algorithm presented by Arruda and coauthors in \cite{arrudaDevelopmentValidationECG1998}.}
\label{tab:arruda}
\end{table}

\end{document}